\documentclass{CVM}
\usepackage{graphicx}
\usepackage{amsmath}
\usepackage{amssymb}
\usepackage{multirow}
\usepackage{mathrsfs}
\usepackage{tabulary}
\usepackage{booktabs}
\usepackage{silence}
\usepackage{colortbl}
\usepackage{xcolor} 
\usepackage{tabularx}
\usepackage{array}
\usepackage{float}
\usepackage{capt-of}
\usepackage{pifont}  %
\usepackage{subcaption} %

\usepackage[bookmarks=false,colorlinks=true,linkcolor=black,citecolor=black,filecolor=black,urlcolor=black]{hyperref}
\usepackage{cleveref}
\usepackage{url}
\usepackage{tikz}

\newcolumntype{x}[1]{>{\centering\arraybackslash}p{#1pt}}

\newlength\savewidth

\newcommand{\PAR}[1]{\vskip4pt \noindent{\bf #1~}}

\newcommand{\tocite}[1]{\textcolor{red}{[TOCITE]}}

\definecolor{cxcolor}{RGB}{  0, 204, 153}
\definecolor{xtcolor}{RGB}{255, 140,   0}

\definecolor{custompink}{RGB}{255, 204, 204} %
\definecolor{customblue}{RGB}{179, 224, 255} %
\definecolor{tablecolor}{rgb}{0,0,0}

\CVMsetup{
type      = {Review Article},
doi       = {CVM.XXXX},
title     = {Towards Depth Foundation Model: Recent Trends in Vision-Based Depth Estimation},
author    = {Zhen Xu$^{*,1}$, Hongyu Zhou$^{*,1}$, Sida Peng$^{1}$, Haotong Lin$^{1}$, Haoyu Guo$^{2}$, Jiahao Shao$^{1}$, Peishan Yang$^{1}$, Qinglin Yang$^{1}$, Sheng Miao$^{1}$, Xingyi He$^{1}$, Yifan Wang$^{1}$, Yue Wang$^{1}$, Ruizhen Hu$^{3}$, Yiyi Liao$^{1}$, Xiaowei Zhou$^{1}$, Hujun Bao$^{1}$\cor{}},
runauthor = {Z. Xu, H. Zhou et al.},
abstract  = {
  Depth estimation is a fundamental task in 3D computer vision, crucial for applications such as 3D reconstruction, free-viewpoint rendering, robotics, autonomous driving, and AR/VR technologies.
  Traditional methods relying on hardware sensors like LiDAR are often limited by high costs, low resolution, and environmental sensitivity, limiting their applicability in real-world scenarios.
  Recent advances in vision-based methods offer a promising alternative, yet they face challenges in generalization and stability due to either the low-capacity model architectures or the reliance on domain-specific and small-scale datasets.
  The emergence of scaling laws and foundation models in other domains has inspired the development of "depth foundation models"—deep neural networks trained on large datasets with strong zero-shot generalization capabilities.
  This paper surveys the evolution of deep learning architectures and paradigms for depth estimation across the monocular, stereo, multi-view, and monocular video settings.
  We explore the potential of these models to address existing challenges and provide a comprehensive overview of large-scale datasets that can facilitate their development.
  By identifying key architectures and training strategies, we aim to highlight the path towards robust depth foundation models, offering insights into their future research and applications.
},
keywords  = {Depth Estimation, Foundation Models, 3D Computer Vision},
copyright = {Zhejiang University},
}

\begin{document}

\maketitle

\enlargethispage{-3pt}
\begin{figure}[b] \vskip -4mm
    \small\renewcommand\arraystretch{1.3}
    \begin{tabular}{p{80.5mm}} \toprule\\ \end{tabular}
    \vskip -4.5mm \noindent \setlength{\tabcolsep}{1pt}
    \begin{tabular}{p{3.5mm}p{80mm}}
        $^*$      & The first two authors contributed equally to this work.                                                                                                                                                                                                                                                                                                          \\
        $1\quad $ & Zhejiang University, China.                                                                                                                                                                                                                                                                                                                                      \\
        $2\quad $ & Shanghai AI Lab, China.                                                                                                                                                                                                                                                                                                                                          \\
        $3\quad $ & Shenzhen University, China.                                                                                                                                                                                                                                                                                                                                      \\
                  & Authors' emails: zhenx@zju.edu.cn, hongy\_zhou@zju.edu.cn, pengsida@zju.edu.cn, haotongl@outlook.com, guohaoyu@pjlab.org.cn, jhshao@zju.edu.cn, yangps0306@gmail.com, yangqinglin@zju.edu.cn, 12231099@zju.edu.cn, hexingyi8@gmail.com, accwyf@gmail.com, ywang24@zju.edu.cn, ruizhen.hu@szu.edu.cn, yiyi.liao@zju.edu.cn, xwzhou@zju.edu.cn, bao@cad.zju.edu.cn \\
                  & \hspace{-5mm} Manuscript received: 2025-01-01; accepted: 2025-01-01\vspace{-2mm}
    \end{tabular} \vspace {-3mm}
\end{figure}

\section{Introduction}
\label{sec:introduction}

\begin{figure*}[ht]
    \centering
    \includegraphics[width=\linewidth]{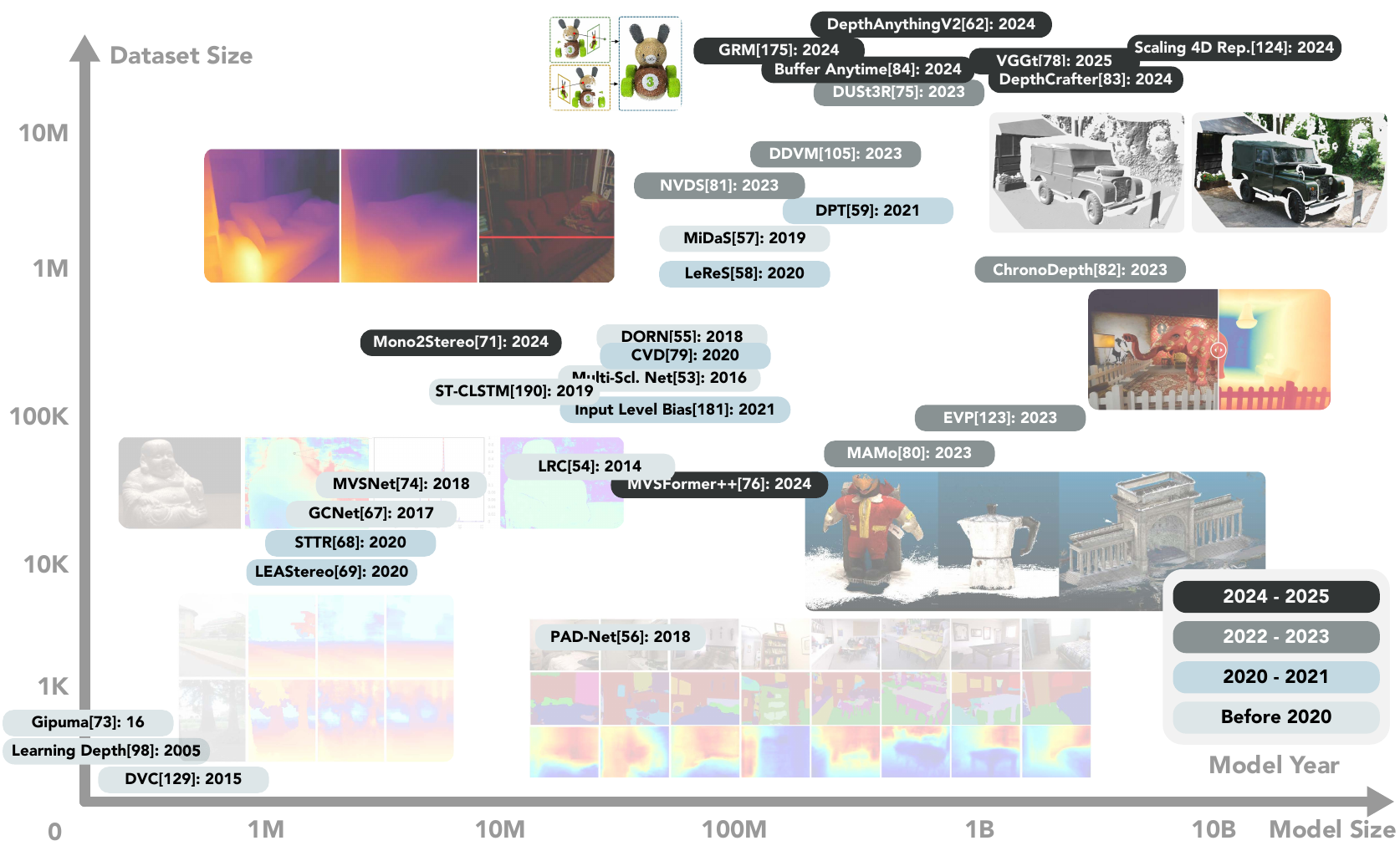}
    \caption{\textbf{Scaling trends in model capacity and data volume for depth estimation.}
        Each point represents a published method, positioned by its approximate model size (bottom axis, logarithmic scale of parameter count) and dataset size (left axis, logarithmic scale of training images), and colored by publication year. Early models (lightest shading, before 2020) relied on sub-million-parameter networks trained on only thousands of images, yielding limited generalization. Over 2020–2021 (light blue shading), methods grew to tens of millions of parameters and hundreds of thousands of image datasets. 
        In 2022-2023 (gray shading), methods continued to increase in size and scalable training, with around one billion parameters and datasets ranging from several million to ten million images. The most recent models (darkest shading, 2024–2025) further scale up to the billions of parameters and utilize over ten million images, with strong generalization ability, and demonstrate the potential evolution toward depth foundation models. 
    }
    \label{fig:overview}
\end{figure*}

\definecolor{leafgreen}{rgb}{0.05, 0.54, 0.25}
\definecolor{lightorange}{RGB}{255, 140, 66}
\definecolor{lightpurple}{RGB}{157, 136, 197}
\definecolor{lightblue}{RGB}{115, 186, 215}

\definecolor{coralpink}{RGB}{255, 180, 190}
\definecolor{dustyblue}{RGB}{138, 158, 168}
\definecolor{lightyellow}{RGB}{255, 215, 0}

\newcommand{\mono}{\color{leafgreen}{\ding{52}}} %
\newcommand{\stereo}{\color{lightorange}{\ding{52}}} %
\newcommand{\multi}{\color{lightpurple}{\ding{52}}} %
\newcommand{\video}{\color{lightblue}{\ding{52}}} %

\newcommand{\bwd}{3.5mm}
\newcommand{\bgap}{\hspace{0.6mm}}

\newcommand{\object}{\textcolor{coralpink}{\rule{\bwd}{\bwd}}}
\newcommand{\indoor}{\textcolor{dustyblue}{\rule{\bwd}{\bwd}}}
\newcommand{\outdoor}{\textcolor{lightyellow}{\rule{\bwd}{\bwd}}}

\begin{table*}
    \centering
    \renewcommand{\arraystretch}{1.0}
    \resizebox{\textwidth}{!}{%
        \begin{tabular}{c | c c c c | c c c c c c c c c}

            \toprule

            \multirow{2}{*}{\textbf{Domain}} & \multicolumn{4}{c|}{\textbf{Tasks}} & \multirow{2}{*}{\textbf{Dataset Name}} & \multirow{2}{*}{\textbf{Dynamic}} & \multirow{2}{*}{\textbf{Metric}} & \multirow{2}{*}{\textbf{Scenes\#}}                                       & \multirow{2}{*}{\textbf{Frames\#}} & \multirow{2}{*}{\textbf{Resolution}} & \multirow{2}{*}{\textbf{Anno}} & \multirow{2}{*}{\textbf{License}}                                               \\
                                             & \small{Mono}                        & \small{Stereo}                         & \small{MV}                        & \small{Video}                    &                                                                          &                                    &                                      &                                &                                   &                  &        &                 \\
            \midrule

            \multirow{45}{*}{\rotatebox{90}{\textbf{~~~~~~~~~~~~~~~~~~~~~~~~~~~~~~~~~~~~~~~~~~~~~~~~~~~~~~Real-World}}}
                                             & \mono                               &                                        &                                   &                                  & A2D2~\cite{geyer2020a2d2} \outdoor                                       & Dynamic                            & Yes                                  & 3                              & 394K                              & $1920\times1208$ & Sparse & CC BY-ND 4.0    \\
                                             & \mono                               & \stereo                                &                                   &                                  & Arogoverse2~\cite{Argoverse2,TrustButVerify} \outdoor                    & Dynamic                            & Yes                                  & 1000                           & 2.14M                             & $2048\times1550$ & Sparse & CC BY-NC-SA 4.0 \\
                                             & \mono                               & \stereo                                &                                   &                                  & CityScapes~\cite{cordts2016cityscapes} \outdoor                          & Dynamic                            & Yes                                  & 2975                           & 89K                               & $2048\times1024$ & Sparse & Non-Commercial  \\
                                             & \mono                               &                                        &                                   &                                  & DDAD~\cite{packnet} \outdoor                                             & Dynamic                            & Yes                                  & 200                            & 98K                               & $1936\times1216$ & Sparse & CC BY-NC-SA 4.0 \\
                                             & \mono                               & \stereo                                &                                   &                                  & DIML~\cite{cho2021diml} \indoor \bgap \outdoor                           & Static                             & Yes                                  & 200+                           & 2M                                & $1334\times756$  & Sparse & Non-Commercial  \\
                                             & \mono                               &                                        &                                   &                                  & DIODE~\cite{vasiljevic2019diode} \indoor \bgap \outdoor                  & Static                             & Yes                                  & 30                             & 27K                               & $1024\times768$  & Sparse & MIT             \\
                                             & \mono                               & \stereo                                &                                   &                                  & DSEC~\cite{Gehrig21ral} \outdoor                                         & Dynamic                            & Yes                                  & 53                             & 26K                               & $1440\times1080$ & Sparse & CC BY-NC-SA 4.0 \\
                                             & \mono                               &                                        &                                   &                                  & HM3D~\cite{ramakrishnan2021hm3d} \indoor                                 & Static                             & Yes                                  & 1000                           & \dag                              & \dag             & Dense  & MIT             \\
                                             & \mono                               &                                        &                                   &                                  & iBims-1~\cite{koch2018evaluation} \indoor                                & Static                             & Yes                                  & 20                             & 54                                & $640\times480$   & Sparse & CC BY 4.0       \\
                                             & \mono                               &                                        &                                   &                                  & Lyft~\cite{houston2021one} \outdoor                                      & Dynamic                            & Yes                                  & 366                            & 158K                              & $1224\times1024$ & Sparse & CC0 1.0         \\
                                             & \mono                               &                                        &                                   &                                  & Mapillary PSD~\cite{antequera2020mapillary} \outdoor                     & Dynamic                            & Yes                                  & 50K                            & 750K                              & $1920\times1080$ & Sparse & CC BY-NC-SA     \\
                                             & \mono                               &                                        &                                   &                                  & NuScenes~\cite{caesar2020nuscenes} \outdoor                              & Dynamic                            & Yes                                  & 1K                             & 40K                               & $1600\times900$  & Sparse & Non-Commercial  \\
                                             & \mono                               &                                        &                                   &                                  & NYU~\cite{silberman2012indoor} \indoor                                   & Static                             & Yes                                  & 464                            & 435K                              & $640\times480$   & Dense  & GPL             \\
                                             & \mono                               &                                        &                                   &                                  & Pandaset~\cite{xiao2021pandaset} \outdoor                                & Dynamic                            & Yes                                  & 103                            & 8K                                & $1920\times1080$ & Sparse & CC0             \\
                                             & \mono                               &                                        &                                   &                                  & Taskonomy\cite{zamir2018taskonomy} \indoor                               & Static                             & Yes                                  & 600                            & 4.5M                              & $512\times512$   & Dense  & MIT             \\
                                             & \mono                               & \stereo                                &                                   &                                  & KITTI~\cite{geiger2012we} \outdoor                                       & Dynamic                            & Yes                                  & 22                             & 41K                               & $1242\times375$  & Sparse & CC BY-NC-SA 3.0 \\
                                             & \mono                               & \stereo                                &                                   &                                  & DrivingStereo~\cite{yang2019drivingstereo} \outdoor                      & Dynamic                            & Yes                                  & 184                            & 180K                              & $1762\times800$  & Sparse & MIT             \\
                                             &                                     & \stereo                                &                                   &                                  & InStereo2K~\cite{bao2020instereo2k} \indoor                              & Static                             & Yes                                  & 50                             & 2K                                & $1080\times860$  & Dense  & Non-Commercial  \\
                                             &                                     & \stereo                                & \multi                            &                                  & Argoverse~\cite{chang2019argoverse} \outdoor                             & Dynamic                            & Yes                                  & 113                            & 6,624                             & $2056\times2464$ & Sparse & CC BY-NC-SA 4.0 \\
                                             &                                     & \stereo                                & \multi                            &                                  & ETH3D~\cite{eth3d} \indoor \bgap \outdoor                                & Static                             & Yes                                  & 25                             & 1K                                & $6233\times4146$ & Dense  & CC BY-NC-SA 4.0 \\
                                             &                                     &                                        & \multi                            & \video                           & Waymo~\cite{sun2020scalability} \outdoor                                 & Dynamic                            & Yes                                  & 1,150                          & 160K                              & $1920\times1280$ & Sparse & Non-Commercial  \\
                                             &                                     &                                        & \multi                            & \video                           & UASOL~\cite{Bauer2019UASOLAL} \outdoor                                   & Static                             & Yes                                  & 676                            & 160.9K                            & $2280\times1282$ & Dense  & CC BY-NC-SA 3.0 \\
                                             &                                     &                                        & \multi                            &                                  & DTU~\cite{dtu} \object                                                   & Static                             & Yes                                  & 124                            & 42.5K                             & $1600\times1200$ & Dense  & MIT             \\
                                             &                                     &                                        & \multi                            &                                  & BlendedMVS~\cite{yao2020blendedmvs} \object \bgap \indoor \bgap \outdoor & Static                             & No                                   & 113                            & 17k                               & $2048\times1536$ & Dense  & CC BY 4.0       \\
                                             & \mono                               &                                        &                                   & \video                           & WildRGBD~\cite{xia2024rgbd} \object                                      & Static                             & Yes                                  & 23K                            & 6M                                & $480\times640$   & Dense  & MIT             \\
                                             & \mono                               &                                        &                                   & \video                           & MVImgNet~\cite{Yu2023MVImgNetAL} \object                                 & Static                             & No                                   & 220K                           & 6.8M                              & $1080\times1920$ & Dense  & CC BY-NC 4.0    \\
                                             & \mono                               &                                        &                                   & \video                           & ARKitScenes~\cite{baruch2021arkitscenes} \indoor                         & Static                             & Yes                                  & 5,047                          & 450M                              & $256\times192$   & Dense  & Non-Commercial  \\
                                             & \mono                               &                                        &                                   & \video                           & ARKitScenes-HighRes~\cite{baruch2021arkitscenes} \indoor                 & Static                             & Yes                                  & 5,047                          & 450M                              & $1920\times1440$ & Dense  & Non-Commercial  \\
                                             & \mono                               &                                        &                                   & \video                           & Matterport3D~\cite{chang2017matterport3d} \indoor                        & Static                             & Yes                                  & 90                             & 194.4K                            & $1280\times1024$ & Dense  & MIT             \\
                                             & \mono                               & \stereo                                &                                   &                                  & Replica\cite{replica19arxiv} \indoor                                     & Static                             & Yes                                  & 18                             & 36K                               & $1200\times680$  & Dense  & Non-Commercial  \\
                                             & \mono                               & \stereo                                & \multi                            & \video                           & Dynamic Replica~\cite{karaev2023dynamicstereo} \indoor                   & Dynamic                            & Yes                                  & 484                            & 145K                              & $1280\times720$  & Dense  & CC BY-NC 4.0    \\
                                             & \mono                               &                                        &                                   & \video                           & ScanNet~\cite{dai2017scannet} \indoor                                    & Static                             & Yes                                  & 1,513                          & 2.5M                              & $640\times480$   & Dense  & Non-Commercial  \\
                                             & \mono                               &                                        &                                   & \video                           & ScanNet++~\cite{yeshwanth2023scannet++} \indoor                          & Static                             & Yes                                  & 1,858                          & 3.7M+                             & $1920\times1440$ & Dense  & Non-Commercial  \\
            \midrule
            \multirow{24}{*}{\rotatebox{90}{\textbf{~~~~~~~~~~~~~~~~~~~~~~~~~~~~~Synthetic}}}
                                             & \mono                               &                                        & \multi                            &                                  & Sintel~\cite{butler2012naturalistic} \indoor \bgap \outdoor              & Dynamic                            & No                                   & 10                             & 1K                                & $1024\times436$  & Dense  & CC BY 3.0       \\
                                             &                                     & \stereo                                &                                   &                                  & SceneFlow~\cite{mayer2016large} \object \bgap \indoor \bgap \outdoor     & Dynamic                            & No                                   & 9                              & 40K                               & $960\times540$   & Dense  & CC BY 4.0       \\
                                             &                                     & \stereo                                &                                   &                                  & CREStereo~\cite{li2022practical} \object                                 & Static                             & No                                   & 0                              & 103K                              & $1920\times1080$ & Dense  & Apache-2.0      \\
                                             &                                     & \stereo                                &                                   &                                  & FallingThings~\cite{tremblay2018falling} \indoor \bgap \outdoor          & Dynamic                            & No                                   & 3                              & 62K                               & $960\times540$   & Dense  & CC BY-NC-SA 4.0 \\
                                             &                                     & \stereo                                &                                   &                                  & FSD~\cite{wen2025foundationstereo} \object \bgap \indoor \bgap \outdoor  & Dynamic                            & No                                   & 12                             & 1M                                & $1280\times720$  & Dense  & Non-Commercial  \\
                                             &                                     & \stereo                                &                                   &                                  & UnrealStereo4K~\cite{tosi2021smd} \indoor \bgap \outdoor                 & Static                             & No                                   & 8                              & 7,720                             & $3840\times2160$ & Dense  & MIT             \\
                                             &                                     & \stereo                                & \multi                            & \video                           & Spring~\cite{mehl2023spring}                                             & Dynamic                            & No                                   & 47                             & 6K                                & $1920\times1080$ & Dense  & CC BY 4.0       \\
                                             &                                     & \stereo                                & \multi                            & \video                           & TartanAir~\cite{wang2020tartanair} \indoor \bgap \outdoor                & Static                             & Yes                                  & 163                            & 1M                                & $640\times480$   & Dense  & CC BY 4.0       \\
                                             &                                     & \stereo                                & \multi                            & \video                           & VirtualKITTI2~\cite{cabon2020vkitti2} \outdoor                           & Dynamic                            & Yes                                  & 5                              & 25K                               & $1242\times375$  & Dense  & CC BY-NC-SA 4.0 \\
                                             &                                     &                                        & \multi                            & \video                           & MatrixCity~\cite{Li2023MatrixCityAL} \outdoor                            & Static                             & Yes                                  & 3K                             & 519K                              & $1000\times1000$ & Dense  & Apache-2.0      \\
                                             & \mono                               &                                        & \multi                            & \video                           & MVS-Synth~\cite{huang2018deepmvs} \outdoor                               & Dynamic                            & No                                   & 120                            & 12K                               & $1920\times1080$ & Dense  & Non-Commercial  \\
                                             & \mono                               &                                        & \multi                            &                                  & 3D Ken Burns~\cite{niklaus20193d} \outdoor                               & Static                             & No                                   & 32                             & 536K                              & $512\times512$   & Dense  & CC BY-NC-SA 4.0 \\
                                             & \mono                               &                                        &                                   & \video                           & OmniObject3D~\cite{wu2023omniobject3d} \object                           & Static                             & Yes                                  & 6K                             & 600K                              & $800\times800$   & Dense  & CC BY 4.0       \\
                                             & \mono                               & \stereo                                &                                   & \video                           & IRS~\cite{wang2021irs} \indoor                                           & Static                             & Yes                                  & 70                             & 100K                              & $960\times540$   & Dense  & Apache 2.0      \\
                                             &                                     &                                        & \multi                            & \video                           & PointOdyssey~\cite{zheng2023pointodyssey} \indoor \bgap \outdoor         & Dynamic                            & No                                   & 131                            & 200K                              & $540\times960$   & Dense  & MIT             \\
                                             &                                     &                                        & \multi                            & \video                           & BEDLAM~\cite{black2023bedlam} \object \bgap \outdoor                     & Dynamic                            & Yes                                  & 10,450                         & 380K                              & $1280\times720$  & Dense  & Non-Commercial  \\

            \bottomrule
        \end{tabular}
    }
    \caption{\textbf{Datasets.} The term \object \bgap indicates that the scenes in these datasets are object-centric. Similarly, \indoor \bgap and \outdoor \bgap refer to indoor and outdoor scenes, respectively. The depth annotations can be classified as dense or sparse, depending on whether most pixels have corresponding depth values. \dag: HM3D is a digital twin dataset created from real-world data, with the number of frames and resolutions being user-defined.}
    \label{tab:datasets}
\end{table*}

Depth estimation stands as a cornerstone in the field of 3D computer vision.
This task has been a focal point for researchers due to its critical role in various applications such as 3D reconstruction, 3D generative models, robotics, autonomous driving, and AR/VR technologies.
However, algorithms often struggle to achieve high-quality and consistent depth recovery akin to human perception, which leverages prior knowledge of the scene and the world.
Traditional depth recovery methods typically rely on active sensing hardware, including commercially available LiDAR, time-of-flight (ToF) sensors, and ultrasonic probes.
These sensors estimate depth by measuring the time taken for photons or sound waves to travel back and forth.
Despite their accuracy, the high cost of these sensors limits their widespread application.
Additionally, active sensors often suffer from low resolution and significant noise interference.
For instance, the LiDAR sensor on an iPhone can only achieve a reconstruction resolution within a limited range and struggles with high precision for very close or distant objects.
Moreover, these sensors are sensitive to environmental lighting conditions, making them less effective in outdoor, high-light scenarios.

Recently, there has been a growing interest in vision-based depth estimation methods that avoid active depth-sensing hardware, instead relying on readily available cameras commonly found in everyday devices.
Compared to active sensor-based approaches, vision methods are cost-effective, offer an unrestricted depth range, are less affected by environmental conditions, and provide high resolution.
For example, a standard iPhone camera can easily capture 4K resolution RGB information.
However, existing vision-based depth estimation algorithms still face numerous challenges.
Monocular depth estimation, in particular, is highly ill-posed, and standard deep learning algorithms struggle to achieve high-precision results.
To introduce constraints and reduce ill-posedness, researchers often explore depth estimation algorithms with multiple camera inputs or scenarios with more extensive scene observations, such as stereo, multi-view, or video-based depth estimation.
Yet, these methods are often trained on small-scale synthetic data, leading to instability across spatial and temporal domains, poor generalization across different scenes and input types, and difficulties in overcoming the domain gap between synthetic and real-world data.

With the validation and rise of scaling laws in natural language processing, text-based image generation, and video generation models, the concept of foundation models has emerged.
Foundation models are deep neural networks trained on vast amounts of data, exhibiting emergent zero-shot generalization capabilities in other domains.
To achieve such capabilities, researchers focus on the scale and variation of input training data, leveraging large-scale models from other domains and ingeniously constructing self-supervision architectures.
We define scalable depth estimation models and architectures capable of absorbing massive data as "\textit{depth foundation models}".
For depth estimation tasks, including monocular, stereo, multi-view, and monocular video depth estimation, corresponding foundation models have the potential to address the aforementioned generalization issues and provide key solutions to long-standing Computer Vision tasks.

This paper aims to survey the evolution towards depth foundation models and paradigms for depth estimation across the monocular, stereo, multi-view, and monocular video settings.
\begin{itemize}
    \item We explore the development of deep learning model architectures and learning paradigms for each task and identify key paradigms with foundational capability or potential.
    \item To aid the development of such depth foundation models, we also provide comprehensive surveys on large-scale datasets in each respective subfield.
    \item We also list the current key challenges faced by the foundational architectures in each task to provide insight into future works.
\end{itemize}

\begin{table*}[ht]
	\centering
	\renewcommand{\arraystretch}{1.0}
	\resizebox{\textwidth}{!}{%
		{\color{tablecolor}
				\begin{tabular}{c|l|l|ll|cc|cc}
					\toprule
					\multirow{2}{*}{\rotatebox[origin=c]{90}{\textbf{Task}}} & \textbf{Method}                                                            & \textbf{Year}                                                 & \textbf{Model Type} & \textbf{Key Paradigm} & \textbf{Model Size} & \textbf{Data Size} & \multicolumn{2}{c}{\textbf{Benchmarks}}                                          \\
					\multicolumn{7}{c}{}                                     & NYUv2 \cite{silberman2012indoor} (Rel.) $\downarrow$                          & KITTI \cite{geiger2012we,geiger2012we} (Rel.) $\downarrow$                                                                                                                                                                             \\
					\multirow{13}{*}{\rotatebox[origin=c]{90}{Monocular}}
					                                                         & Multi-Scale Net \cite{eigen2014depth}                                      & 2014                                                          & CNN                 & Direct Regression     & $\sim$70M           & $\sim$150K         & \cellcolor{gray!10}--                         & \cellcolor{gray!10}0.190         \\
					                                                         & LRC \cite{godard2017unsupervised}                                          & 2016                                                          & CNN                 & Direct Regression     & $\sim$30M           & $\sim$50K          & \cellcolor{gray!10}--                         & \cellcolor{gray!10}0.114         \\
					                                                         & DORN \cite{fu2018deep}                                                     & 2018                                                          & CNN                 & Depth Classification  & $\sim$50M           & $\sim$200K         & \cellcolor{gray!10}0.115                      & \cellcolor{gray!10}0.072         \\
					                                                         & PAD-Net \cite{xu2018pad}                                                   & 2018                                                          & CNN                 & Multiple Tasks        & $\sim$25M           & $\sim$3K           & \cellcolor{gray!10}0.120                      & \cellcolor{gray!10}--            \\
					                                                         & MiDaS \cite{ranftl2020towards}                                             & 2019                                                          & CNN                 & Affine Invariant      & $\sim$90M           & $\sim$2M           & \cellcolor{gray!10}0.111                      & \cellcolor{gray!10}0.236         \\
					                                                         & LeReS \cite{yin2021learning}                                               & 2020                                                          & CNN                 & Canonical Camera      & $\sim$90M           & $\sim$350K         & \cellcolor{gray!10}0.090                      & \cellcolor{gray!10}0.149         \\
					                                                         & DPT \cite{ranftl2021vision}                                                & 2021                                                          & ViT                 & Scaling Law           & $\sim$350M          & $\sim$1.5M         & \cellcolor{gray!10}0.098                      & \cellcolor{gray!10}0.100         \\
					                                                         & Marigold \cite{ke2024repurposing}                                          & 2024                                                          & Diffusion           & Scaling Law           & $\sim$800M          & $\sim$75K          & \cellcolor{gray!10}0.055                      & \cellcolor{gray!10}0.099         \\
					                                                         & DA \cite{yang2024depth}                                                    & 2024                                                          & ViT                 & Scaling Law           & $\sim$600M          & $\sim$63.5M        & \cellcolor{gray!30}0.0382                     & \cellcolor{gray!30}0.0504        \\
					                                                         & DAV2 \cite{yang2024depthv2}                                                & 2024                                                          & ViT                 & Scaling Law           & $\sim$2B            & $\sim$62.5M        & \cellcolor{gray!30}0.0416                     & \cellcolor{gray!30}0.0677        \\
					                                                         & MoGE \cite{wang2024moge}                                                   & 2024                                                          & ViT                 & Scaling Law           & $\sim$350M          & $\sim$4M           & \cellcolor{gray!30}0.0292                     & \cellcolor{gray!30}0.0394        \\
					                                                         & UniDepth \cite{piccinelli2024unidepth}                                     & 2024                                                          & ViT                 & Metric Depth          & $\sim$350M          & $\sim$4M           & \cellcolor{gray!30}0.0340/0.116               & \cellcolor{gray!30}0.0350/0.0469 \\
					                                                         & UniDepthV2 \cite{piccinelli2025unidepthv2}                                 & 2025                                                          & ViT                 & Metric Depth          & $\sim$350M          & $\sim$4M           & \cellcolor{gray!30}0.0296/0.106               & \cellcolor{gray!30}0.0385/0.0858 \\
					                                                         & MoGE2 \cite{wang2025moge2accuratemonoculargeometry}                        & 2025                                                          & ViT                 & Metric Depth          & $\sim$350M          & $\sim$4M           & \cellcolor{gray!30}0.0289/0.073               & \cellcolor{gray!30}0.0375/0.1810 \\
					\midrule
					\multicolumn{7}{c}{}                                     & \multicolumn{2}{c}{KITTI 2015 \cite{geiger2012we} (D1) $\downarrow$}                                                                                                                                                                                                                                                  \\
					\multirow{5}{*}{\rotatebox[origin=c]{90}{Stereo}}
					                                                         & GCNet \cite{kendall2017gcnet}                                              & 2017                                                          & CNN                 & Cost Volume           & $\sim$4M            & $\sim$35K          & \multicolumn{2}{c}{\cellcolor{gray!10}2.87\%}                                    \\
					                                                         & STTR \cite{li2021revisitingsstr}                                           & 2020                                                          & ViT                 & Attention             & $\sim$3M            & $\sim$22K          & \multicolumn{2}{c}{\cellcolor{gray!10}2.01\%}                                    \\
					                                                         & LEAStereo \cite{cheng2020hierarchical}                                     & 2020                                                          & CNN                 & Attention             & $\sim$2M            & $\sim$20K          & \multicolumn{2}{c}{\cellcolor{gray!10}1.65\%}                                    \\
					                                                         & RAFT-Stereo \cite{lipson2021raft}                                          & 2021                                                          & CNN+GRU             & Iterative Optim.      & $\sim$10M           & $\sim$35K          & \multicolumn{2}{c}{\cellcolor{gray!10}1.96\%}                                    \\
					                                                         & Mono2Stereo \cite{wang2024mono2stereo}                                     & 2024                                                          & CNN                 & Scaling Law           & $\sim$5M            & $\sim$600K         & \multicolumn{2}{c}{\cellcolor{gray!10}1.58\%}                                    \\
					\midrule
					\multicolumn{7}{c}{}                                     & \multicolumn{2}{c}{DTU \cite{jensen2014large} (Overall Err.) $\downarrow$}                                                                                                                                                                                                                                             \\
					\multirow{6}{*}{\rotatebox[origin=c]{90}{Multi-view}}
					                                                         & Gipuma \cite{galliani2015massively}                                        & 2015                                                          & Handcrafted         & Handcrafted Algo.     & $\sim$0             & 0                  & \multicolumn{2}{c}{\cellcolor{gray!30}0.283}                                     \\
					                                                         & MVSNet \cite{mvsnet}                                                       & 2018                                                          & CNN                 & Cost Volume           & $\sim$4M            & $\sim$27K          & \multicolumn{2}{c}{\cellcolor{gray!30}0.462}                                     \\
					                                                         & DUSt3R \cite{dust3r_cvpr24}                                                & 2023                                                          & ViT                 & Scaling Law           & $\sim$400M          & $\sim$17M          & \multicolumn{2}{c}{\cellcolor{gray!30}1.741}                                     \\
					                                                         & MVSFormer++ \cite{cao2024mvsformer++}                                      & 2024                                                          & ViT                 & Scaling Law           & $\sim$40M           & $\sim$40K          & \multicolumn{2}{c}{\cellcolor{gray!30}0.281}                                     \\
					                                                         & MASt3R \cite{mast3r_arxiv24}                                               & 2024                                                          & ViT                 & Scalable Law          & $\sim$400M          & $\sim$20M          & \multicolumn{2}{c}{\cellcolor{gray!30}0.374}                                     \\
					                                                         & VGGt \cite{wang2025vggt}                                                   & 2025                                                          & ViT                 & Scaling Law           & $\sim$1B            & $\sim$30M          & \multicolumn{2}{c}{\cellcolor{gray!30}0.382}                                     \\
					\midrule
					\multicolumn{7}{c}{}                                     & ScanNet \cite{dai2017scannet} (Rel.) $\downarrow$                          & KITTI \cite{geiger2012we,geiger2012we} (Rel.) $\downarrow$                                                                                                                                                                             \\
					\multirow{6}{*}{\rotatebox[origin=c]{90}{Video}}
					                                                         & CVD \cite{luo2020consistent}                                               & 2020                                                          & CNN                 & Test-Time Optim.      & $\sim$25M           & $\sim$120K         & \cellcolor{gray!10}0.073                      & \cellcolor{gray!10}--            \\
					                                                         & MAMo \cite{yasarla2023mamo}                                                & 2023                                                          & ViT                 & Temporal Corr.        & $\sim$200M          & $\sim$60K          & \cellcolor{gray!30}--                         & \cellcolor{gray!30}0.049         \\
					                                                         & NVDS \cite{wang2023neural}                                                 & 2023                                                          & ViT                 & Scaling Law           & $\sim$100M          & $\sim$2.2M         & \cellcolor{gray!10}0.187                      & \cellcolor{gray!10}0.253         \\
					                                                         & ChronoDepth \cite{shao2024learning}                                        & 2024                                                          & Diffusion           & Scaling Law           & $\sim$2B            & $\sim$240K         & \cellcolor{gray!10}0.159                      & \cellcolor{gray!10}0.151         \\
					                                                         & DepthCrafter \cite{hu2024depthcrafter}                                     & 2024                                                          & Diffusion           & Scaling Law           & $\sim$2B            & $\sim$25M          & \cellcolor{gray!10}0.125                      & \cellcolor{gray!10}0.110         \\
					                                                         & Buffer Anytime \cite{kuang2024buffer}                                      & 2024                                                          & ViT+Diffusion       & Scaling Law           & $\sim$300M          & $\sim$26M          & \cellcolor{gray!10}0.123                      & \cellcolor{gray!10}0.119         \\
					\bottomrule
				\end{tabular}
			}
	}
    \caption{%
			\textbf{Key methods in depth estimation.}
			Model size and data size are approximate. Benchmark results are collected from the corresponding papers. Sections marked with different background colors (\fcolorbox{black}{gray!10}{\strut~} light gray or \fcolorbox{black}{gray!30}{\strut~} medium gray) indicate different evaluation protocols used by the papers.
	}
	\label{tab:key_methods}
\end{table*}

\section{Survey Scope}

This paper primarily concentrates on depth estimation methods that leverage deep learning, with a particular emphasis on foundation models that utilize large-scale architectures and extensive datasets. We begin by defining depth foundation models and then outline the depth estimation tasks that will be addressed in the following sections.

\subsection{Definition of Depth Foundation Models}

We provide a brief overview of the development of foundation models in language models and self-supervised vision-based tasks to facilitate the understanding of the depth of foundation models. The field of language models has experienced explosive growth with the establishment of foundation models in recent years. This progress stems from the ability of these models to learn universal language and patterns from massive datasets, enabling them to generalize powerfully across various downstream tasks.
Moreover, development on scalable architectures and self-supervision tasks has also enabled the emergence of vision foundation models, which are trained on a vast amount of image or video data to facilitate the 2D or 3D perception task.

Convolutional neural networks and long short-term memory networks \cite{graves2012long} played a main role at the early stage of language models, with limited network and data scales.
The concept of Word Embeddings \cite{almeida2019word} and the introduction of the self-attention Mechanism \cite{waswani2017attention} allowed the model to process all words in a sequence simultaneously, vastly improving parallel computation efficiency and the ability to capture long-range dependencies. The original Transformer model had a relatively small number of parameters, but its architecture laid the groundwork for subsequent large-scale models.
BERTs \cite{devlin2019bert} and GPTs \cite{brown2020language} can be considered as the beginning of foundation models in large language models (LLMs).
Proposed by Google, BERT is a bidirectional pre-trained model based on the Transformer architecture, enabling better understanding of the polysemy of words in a sentence. Bert is trained on Toronto BookCorpus (800 million words) and English Wikipedia (2.5 billion words). BERT-Base has 110 million parameters, and BERT-Large has 340 million parameters.
Proposed by OpenAI, GPT is a unidirectional generative pre-trained model based on the Transformer architecture. GPT models learn language patterns by predicting the next word, excelling in text generation tasks. The GPT-3 is trained on a dataset that is larger than 45 TB, along with 175 billion parameters.

Vision foundation models have recently emerged as powerful tools for a wide range of visual perception tasks, closely following the development of scalable model architectures in language models. These models, such as DINO~\cite{caron2021emerging, oquab2023dinov2, simeoni2025dinov3}, MAE~\cite{he2022masked}, and SAM~\cite{kirillov2023segment,ravi2024sam}, are typically trained on massive image or video datasets using self-supervised or weakly supervised objectives. By leveraging scalable architectures (e.g., Vision Transformers) and large-scale pretraining, vision foundation models learn rich and generalizable visual representations that can be transferred to downstream tasks, including classification, detection, segmentation, and even 3D perception. Notably, recent diffusion-based vision models~\cite{rombach2022high, ho2020denoising} have also demonstrated strong capabilities in generative and dense prediction tasks. These advances in vision foundation models provide the architectural and data-driven basis for the development of depth foundation models, enabling improved generalization and scalability for depth estimation across diverse domains.

The development of depth estimation models is illustrated in \cref{fig:overview}. Considering the foundation models scale in the areas of language models, we define a depth foundation model as one that is trained on a large-scale dataset (over 10 million images) and employs models with a substantial number of parameters (over 1 billion). Additionally, depth foundation models should exhibit strong generalizability across multiple data domains.

\subsection{Depth Estimation Tasks}
This survey covers several tasks, including monocular depth estimation, stereo depth estimation, multi-view depth estimation, and monocular video depth estimation using foundation models.
Let $\textbf{I} = \{I_{k,t}, k=1,...,\mathcal{K}, t=1,...,\mathcal{T}\}$ represent a collection of RGB images, where $\mathcal{K}$ denotes the number of cameras and $\mathcal{T}$ is the number of timestamps for the frames.
In the case of monocular depth estimation, the input consists of a single image $I_{1,1}$.
For stereo depth estimation, the input comprises a pair of images $\{I_{1,1}, I_{2,1}\}$. In multi-view depth estimation, the input is a set of images captured at the same timestamp but varying in spatial locations, represented as $\{I_{k,1}, k=1,...,\mathcal{K}\}$.
For monocular video depth estimation, the input consists of a sequence of images captured by a monocular camera at different timestamps, represented as $\{I_{1,t}, t=1,...,\mathcal{T}\}$.
The scope of our survey excludes the task of multi-view video depth estimation, which can be represented as the most general form of inputs: $\{I_{k,t}, k=1,...,\mathcal{K}, t=1,...,\mathcal{T}\}$. This is due to the fact that foundation models for this task have not yet been thoroughly explored.

Moreover, we additionally define the task of metric depth estimation as one that requires the model to not only output the relative depth map, but also predict its real-world scale and offset. For the tasks of stereo, multi-view, and monocular video depth estimation, depth estimation requires predicting depth maps that, when unprojected to the world coordinate system, align across different views. When the real-world metric-scale camera parameters are known, these three tasks automatically become metric depth estimation tasks. Thus, we only distinguish between relative and metric depth estimation in the monocular depth estimation task.

For each task, we begin by reviewing the background and evolution of deep learning models specific to the task. We then delve into the development of foundation models.
Prominent examples of foundation models include transformer-based models and diffusion models. Furthermore, we also discuss the large-scale datasets used for training these foundation models, encompassing both synthetic and real-world datasets, which enable the models to generalize effectively across diverse scenes. Finally, we address valuable problems faced by existing depth foundation models.

\label{sec:scope}

\section{Overview of Depth Estimation}

In this section, we provide an overlook of paradigms and datasets used in depth models.

\PAR{Paradigms.}
In monocular image depth estimation, models have progressed from direct depth regression to affine-invariant depth, depth classification, and canonical camera depth. This progression facilitates more accurate depth map predictions using just a single input image.
In stereo image depth estimation, by utilizing the principles of stereo geometry, models can concentrate on matching corresponding pixels in image pairs. This has driven the evolution of paradigms from cost-volume methods to the attention mechanism, iterative optimizers, and ultimately to scalable training approaches.
In multi-view image depth estimation, similar to stereo paradigms, the incorporation of multi-view information has facilitated an evolution from patch-match stereo to cost volume methods, followed by a coarse-to-fine strategy and the implementation of token attention mechanisms.
In monocular image depth estimation, by incorporating an additional dimension of timestamps, models leverage temporal correlation and test-time optimization paradigms to establish the relationship between temporal and spatial information. Additionally, scaling up training enhances depth estimation performance and guides models toward becoming depth foundation models.

\PAR{Datasets.} We summarize the datasets commonly used in depth estimation tasks in \cref{tab:datasets}. These datasets can be categorized into two classes: real-world captured and synthesized. Each dataset may be applicable to multiple tasks, and we specify the tasks associated with each dataset in the table. Furthermore, we present details about the scenes within the datasets, including whether they provide metric information, if they include dynamic scenes, the number of scenes and frames, the image resolution, and whether the annotations are dense, based on the proportion of pixels that have corresponding depth values.

\label{sec:overview}

\section{Monocular Image Depth Estimation}
\label{sec:monocular_image}

\begin{figure}[ht]
  \centering
  \includegraphics[width=\linewidth]{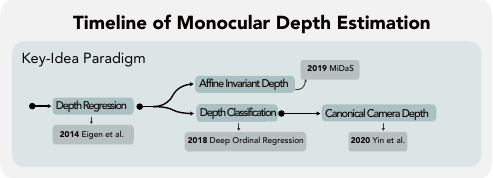}
  \caption{\textbf{Overview of the key-idea paradigm evolution of monocular image depth estimation}.
    From early direct regression and classification methods, through affine-invariant and canonical camera depth estimation, monocular depth estimation models have shown increasingly stronger generalization capabilities, paving the way for the emergence of depth foundation models.
  }
  \label{fig:mono_timeline}
\end{figure}

\begin{figure*}[ht]
  \centering
  \includegraphics[width=\linewidth]{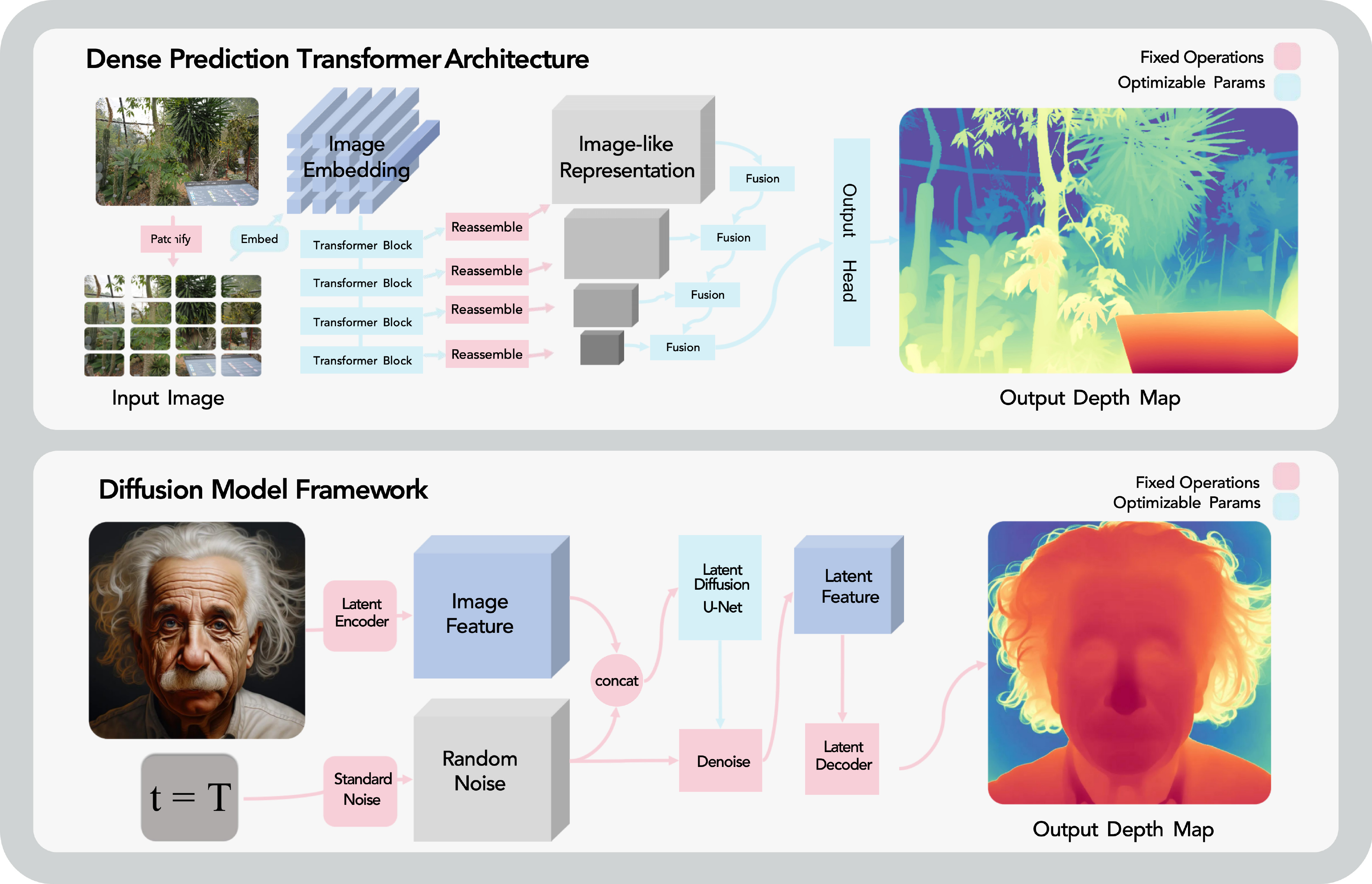}
  \caption{
    \textbf{Representative pipelines of monocular image depth estimation.}
    The \colorbox{custompink}{\strut pink component} denotes operations without learnable parameters or with fixed parameters, while the \colorbox{customblue}{\strut blue component} indicates operations with optimizable parameters.
    Vision Transformer-based approaches, leveraging their lightweight architectures, enable real-time monocular depth estimation.
    However, due to the presence of convolutional operations in their architectures, they may lose detailed features.
    Diffusion Model-based methods treat RGB images as conditional inputs, effectively preserving fine-grained details. Nevertheless, their denoising processes impose computational costs, making it challenging to achieve real-time performance.}
  \label{fig:mono_arch}
\end{figure*}

Monocular depth estimation aims to predict per-pixel depth distances for a scene from a single RGB image, establishing a geometric mapping from 2D images to 3D scenes.
The core challenges lie in scale ambiguity and the lack of geometric information inherent in monocular vision.
Compared to active depth sensors, monocular depth estimation requires no additional hardware and relies solely on image content to infer scene structure, making it valuable for various applications such as 3D reconstruction, video editing, autonomous driving, and augmented reality.
In recent years, with the advancement of deep learning techniques, monocular depth estimation has gradually shifted from traditional geometric methods to data-driven end-to-end learning paradigms, evolving toward foundation models with strong generalization capabilities, high accuracy, and robustness.

\PAR{Evolution of model architectures.} The evolution of monocular depth estimation has been driven by progressive innovations across network architectures, depth representations, and learning paradigms. Early approaches predominantly employed handcrafted image processing pipelines~\cite{saxena2005learning,hoiem2007recovering,liu2008sift,saxena2008make3d}, with seminal works like CNN~\cite{eigen2014depth} establishing the deep learning foundation. Subsequent refinements incorporated U-Net~\cite{eigen2015predicting} and ResNet~\cite{laina2016deeper} residual blocks to enhance spatial continuity. The advent of dense prediction transformers~\cite{ranftl2021vision} marked a paradigm shift: DPT~\cite{ranftl2021vision} introduced global attention mechanisms through patch-wise sequence modeling, addressing CNN-inherent locality constraints. The latest frontier involves diffusion models~\cite{ji2023ddp,saxena2023surprising}, exemplified by Marigold~\cite{ke2024repurposing}, which leverage conditional denoising frameworks to transfer generative priors into depth estimation, significantly improving geometric consistency and cross-domain generalization.

\PAR{Evolution of method paradigms.} Parallel advancements emerged in depth representation and learning strategies. Early methods~\cite{eigen2014depth,eigen2015predicting,laina2016deeper} focused on absolute depth prediction but faced scale ambiguity challenges, prompting innovations like scale-invariant loss~\cite{ranftl2020towards,yin2020diversedepth} for relative depth estimation.
Another advancement came through bin classification approaches~\cite{fu2018deep,lee2019monocular,bhat2021adabins,bhat2022localbins,zhang2023improving,shao2023iebins}, which discretize the continuous depth space into multiple bins. This reformulates depth estimation as a per-pixel classification problem, effectively handling non-uniform depth distributions and capturing uncertainty in predictions, particularly at depth discontinuities.
Recent breakthroughs~\cite{yin2021learning,yin2023metric3d,hu2024metric3d,piccinelli2024unidepth,bochkovskii2024depth} integrate camera parameters to resolve metric ambiguity while maintaining scale awareness. The data paradigm evolved through large-scale multi-view stereo (MVS) datasets like MegaDepth~\cite{li2018megadepth}, enabling unprecedented generalization. Supervision methods expanded beyond fully labeled training: self-supervised approaches exploited photometric consistency in stereo sequences~\cite{godard2019digging}, while Depth Anything~\cite{yang2024depth} demonstrated pseudo-label distillation's effectiveness for knowledge transfer. Multi-task frameworks~\cite{qi2018geonet,xu2018pad,fu2024geowizard,eftekhar2021omnidata} further enhanced robustness through joint depth-flow-pose estimation with geometric constraints, complemented by geometry-aware losses like Virtual Normal that explicitly enforce surface regularity. This multifaceted progression underscores how architectural innovation, representation learning, and supervision paradigms collectively advance monocular depth estimation toward human-level scene understanding.

\PAR{Evolution towards foundation models.} Models for monocular depth estimation are becoming larger and more data-intensive, evolving towards foundation models.
In terms of \textit{Model Size}, advanced foundation models primarily utilize two key methods: Vision Transformer (ViT) architectures \cite{dosovitskiy2020image} and diffusion-based generative models \cite{ho2020denoising}. Dense Prediction Transformers (DPT) are the main architecture for modern depth estimation \cite{ranftl2021vision}, featuring 343 million parameters. Unlike older fully convolutional networks, DPT's ViT backbone keeps high-resolution features and a global view throughout, leading to more detailed and consistent depth estimates. Pre-Trained Diffusion models are used as strong depth estimators to improve how well monocular depth estimation works \cite{song2025depthmaster,ke2024repurposing,xu2024matters,lavreniuk2024evp}, thanks to their existing visual knowledge, which ranges from 200 million to 1 billion parameters. For example, Marigold \cite{ke2024repurposing} modifies text-to-image latent diffusion models (like Stable Diffusion v2 \cite{rombach2022high}) to predict depth from images.
In terms of \textit{Data scale}, powerful hardware and better cameras have led to many high-quality depth datasets, typically ranging from thousands to millions of images; further details can be found in \cref{tab:datasets}. The current trend is to train models on large datasets to make them more adaptable. When training for large-scale depth estimation, models typically use loss functions that ignore scale and shift differences in the data \cite{ranftl2020towards}. This means models learn to predict relative depth, making relative depth estimation a standard task in computer vision. However, metric depth estimation, which provides absolute distance measurements, is crucial for real-world applications. Currently, there are two main ways to get metric depth: 1) Methods like those in \cite{yin2023metric3d, hu2024metric3d,bochkovskii2024depth,piccinelli2024unidepth,wang2024moge,wang2025moge2accuratemonoculargeometry,carreira2024scaling} combine camera intrinsic estimations with relative depth predictions to get metric-scaled outputs through geometric calculations. 2) Directly learning metric depth: Approaches such as those in \cite{bhat2023zoedepthzeroshottransfercombining,viola2024marigolddc,lin2024promptda} train models directly on data that includes scale information, allowing them to predict metric depth without extra steps.

\PAR{Valuable problems.} Recent methods have made notable progress, yet critical challenges persist across four primary dimensions.
\textit{Depth accuracy.}
Depth accuracy enhancement remains the foremost pursuit, aiming to develop a "visual LiDAR" system where RGB cameras rival dedicated depth sensors. Current methods exhibit 4-5\% relative error on standard benchmarks \cite{piccinelli2024unidepth}, escalating to 20-50\% in challenging scenarios \cite{yang2024videodepth}, while hardware sensors consistently achieve sub-1\% accuracy.
\textit{Absolute scale recovery.}
Though improved through works like Metric3D \cite{hu2024metric3d}, it demonstrates fragility under complex illumination and texture-deficient conditions, necessitating more robust geometric priors. The \textit{Data Efficiency Bottleneck} manifests through compounded limitations: pseudo-label noise restricts supervision quality while synthetic-to-real domain gaps constrain model generalization, demanding innovative low-cost high-precision annotation paradigms.
\textit{Multi-task generalization.}
This presents an open research frontier, as current approaches struggle to unify depth estimation with complementary tasks like semantic segmentation and surface normal prediction within a single foundational model architecture. These interconnected challenges collectively underscore the need for fundamental breakthroughs in geometric understanding, data utilization, and cross-task knowledge integration to bridge the performance gap between learning-based methods and physical sensing systems.

\section{Stereo Image Depth Estimation}
\label{sec:stereo_image}

\begin{figure}[ht]
  \centering
  \includegraphics[width=\linewidth]{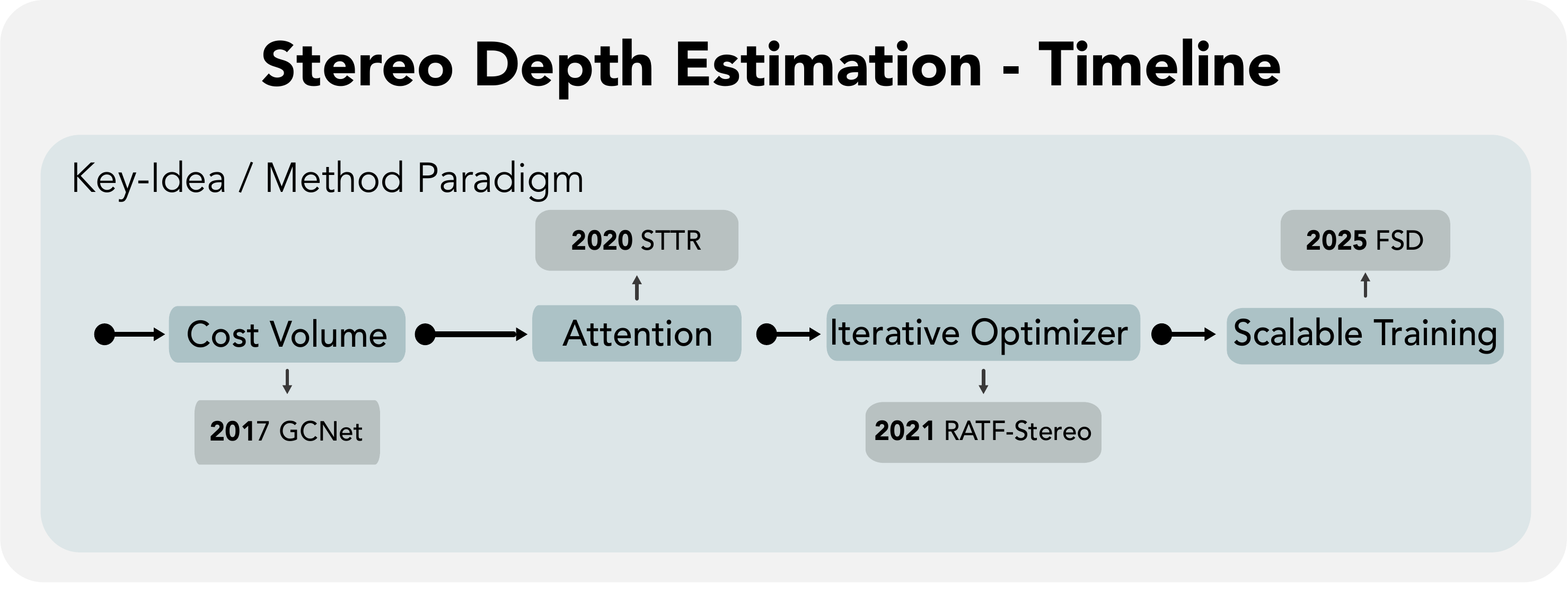}
  \caption{\textbf{Overview of the key-idea paradigm evolution of stereo image depth estimation}.
    The paradigms have transitioned from cost-volume methods to attention mechanisms and iterative optimization techniques to effectively match the features of stereo images.
    The incorporation of monocular and diffusion priors facilitates large-scale training, paving the way for foundation models.
  }
  \label{fig:stereo_timeline}
\end{figure}

\begin{figure*}[ht]
  \centering
  \includegraphics[width=\linewidth]{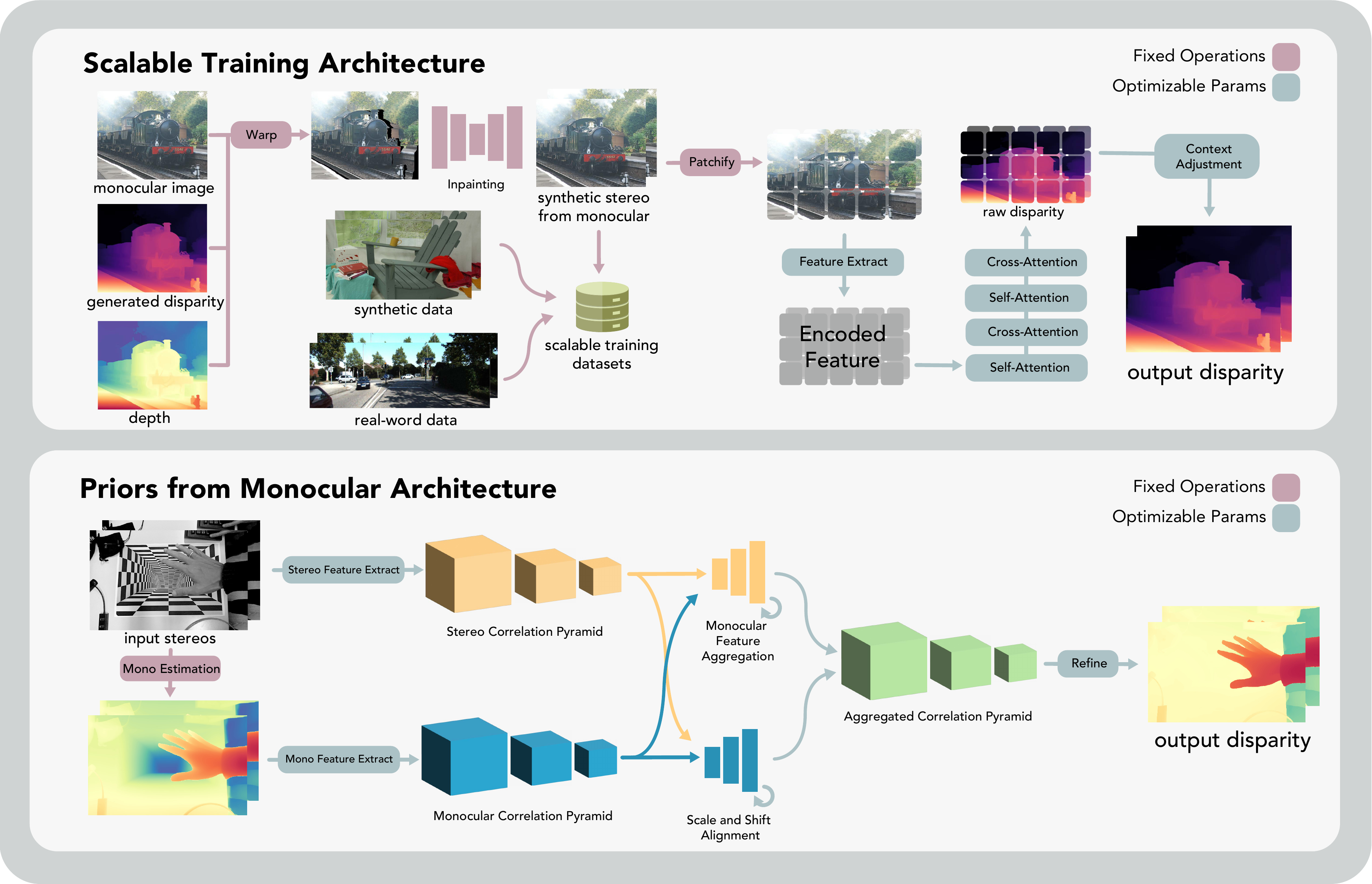}
  \caption{\textbf{Representative pipelines of stereo image depth estimation}.
    The first architecture is for scalable training, which leverages all available datasets along with pseudo stereo pairs synthesized from a monocular dataset, to train a foundation model.
    The second architecture is migrating knowledge from a monocular foundation model to the stereo model, making it possible to achieve a stereo foundation model from relatively small-scale training datasets.
  }
  \label{fig:stereo_arch}
\end{figure*}

Stereo Depth Estimation aims to estimate the per-pixel depth for a scene given the relative pose of a pair of stereo cameras and a pair of RGB observations as input.
Compared to monocular depth estimation, stereo depth estimation can utilize disparity priors to estimate depth through epipolar feature matching and triangulation, which theoretically yields better results than monocular depth estimation.
Additionally, since stereo depth estimation incorporates the known relative pose of the cameras, the estimated depth inherently includes scale information, addressing the scale ambiguity present in monocular depth estimation.
The core challenge of stereo depth estimation lies in accurately establishing the correspondence between pixels in the left and right images. The traditional matching process can be easily affected by factors such as changes in lighting, occlusions, repetitive textures, and weak textures.

With the rapid development of deep learning technology, stereo depth estimation has gradually become popular for using neural networks to replace traditional epipolar feature matching methods. The neural network architecture paradigms mainly include the CNN-based cost-volume paradigm, the Transformer-based attention paradigm, and the RNN-based iterative optimization paradigm.

\PAR{Evolution of model architectures.}
The CNN-based models are one of the most classic and widely used technical approaches in stereo depth estimation \cite{chen2015deep,kendall2017gcnet, chang2018pyramid, guo2019gwcnet, zhang2019ga, tosi2021smd, shen2022pcw, cheng2020hierarchical,badki2020bi3d,xu2020aanet,yang2020waveletstereo}. Early methods, such as PSMNet \cite{chang2018pyramid}, used simple convolution layers to extract features, while subsequent works adopted more complex architectures like ResNet \cite{resnet}, DenseNet \cite{huang2017denselynet}, and NAS search techniques \cite{cheng2020hierarchical} to extract more robust features.
Since 2020, with the success of Transformers \cite{waswani2017attention} in the field of language models and the advancements of Vision Transformers \cite{dosovitskiy2020image} in computer vision, the attention paradigm was proposed \cite{li2021revisitingsstr, guo2022cstr, liu2024goat,xu2023unifying,su2022chitransformer,croco,lou2023elfnet}. To reduce computational complexity, \cite{guo2022cstr} introduces local attention or sparse attention mechanisms.
Inspired by RNNs \cite{medsker2001recurrent} and the application of the RAFT network architecture in optical flow estimation, \cite{lipson2021raft, xu2023iterative, xu2024igev++} construct a correlation pyramid and use an RNN model to iteratively optimize the disparity map. \cite{xu2024igev++} introduces learnable update strategies that dynamically adjust the update direction through attention mechanisms.
\cite{bartolomei2024stereo, cheng2025monster, jiang2025defom, wen2025foundationstereo} and \cite{wang2025stereogen} utilize monocular foundation models as priors and diffusion models as inpainting tools for stereo generation, respectively, advancing stereo depth estimation through large-scale training.

\PAR{Evolution of model paradigms.}
The CNN-based Cost Volume paradigm \cite{kendall2017gcnet, chang2018pyramid, guo2019gwcnet, zhang2019ga, tosi2021smd, shen2022pcw, cheng2020hierarchical,badki2020bi3d,xu2020aanet,yang2020waveletstereo} constructs a cost volume using 2D or 3D CNNs to represent the matching cost between the left and right images, and then performs Cost Aggregation as a post-processing step to ultimately regress the disparity map.
The attention paradigm \cite{li2021revisitingsstr, guo2022cstr, liu2024goat,xu2023unifying,su2022chitransformer,croco,lou2023elfnet, xu2024igev++} models the epipolar matching problem as a sequence-to-sequence task, utilizing Self-Attention and Cross-Attention mechanisms to capture long-range dependencies between pixels.
Within a single image, the self-attention mechanism models the relationships between pixels to capture contextual information, while the Cross-Attention mechanism matches corresponding pixels between the left and right images. Specifically, each pixel in the left image interacts with all pixels in the right image to compute attention weights.
The iterative optimization paradigm \cite{lipson2021raft, xu2023iterative, xu2024igev++} dynamically updates the Cost Volume or feature maps based on the current disparity map to capture more accurate matching information. It can balance performance and speed through early stopping, making it suitable for time-sensitive applications. Additionally, this paradigm exhibits strong robustness to initial errors.
With significant advancements in simulation technology, generative techniques, and monocular depth estimation methods in recent years, foundation models \cite{guo2024stereo, wang2025stereogen, wen2025foundationstereo, wang2020tartanair, watson2020learning, cheng2025monster, jiang2025defom} in the field of stereo depth estimation are beginning to emerge.

\PAR{Evolution towards foundation models.}
Foundation models are emerging as the new paradigm for stereo depth estimation, leading to an increase in data intensity.
However, in contrast to foundation models used for monocular depth estimation, the \textit{Model Size} for stereo depth estimation does not see a significant increase, ranging from 3.5 million to 11 million parameters \cite{kendall2017gcnet, li2021revisitingsstr, cheng2020hierarchical, lipson2021raft}. Instead, advances in monocular depth estimation foundation models allow stereo tasks to leverage monocular priors to enhance depth estimation. There are currently two main approaches to utilizing these priors. One approach \cite{bartolomei2024stereo, wen2025foundationstereo} involves injecting features from monocular depth models into the cost volume. The other approach \cite{cheng2025monster, jiang2025defom} applies stereo metrics to scale monocular depth estimates and then uses refinement networks to achieve better depth estimation results.
There are two methods to enable \textit{Large Data Scale Training} for stereo depth estimation. 1) creating high-fidelity virtual scenes using advanced simulation and rendering technologies, such as FSD \cite{wen2025foundationstereo} (1 million image pairs) and TartanAir \cite{wang2020tartanair} (1 million image pairs). 2) synthesizing stereo data from monocular data. Learning stereo from single images \cite{watson2020learning} (597 thousand image pairs) and Stereo anything\cite{guo2024stereo} (30 million image pairs) generate virtual stereo pairs by using estimated scene depth from monocular images \cite{ranftl2020towards, yang2024depth}, allowing pixels to be warped to pseudo-stereo viewpoints and inpainting to fill in gaps. Recent advancements in diffusion models have led to stereoGen \cite{wang2025stereogen} (35 thousand image pairs) using stable diffusion as an inpainting tool.

\PAR{Valuable Problems.}
Research on foundation models for stereo depth estimation is still in its early stages, and we believe there are several key issues worth exploring in the future:
\textit{Limited data.}
Generated stereo data from monocular faces challenges such as insufficient accuracy in monocular depth estimation and difficulties in filling in warp holes. Additionally, there remains a domain gap between synthetic data and the real world, and the diversity of the synthetic datasets is still not rich enough.
\textit{Lack of end-to-end training paradigm.}
Current methods that leverage monocular priors treat monocular foundation models as cues, lacking end-to-end training of large model parameters. The model parameter counts are relatively small, and there is a lack of foundation model designs tailored for stereo tasks.
\textit{Limit utilization of available datasets.}
There is an insufficient application of cross-domain datasets. Aside from Stereo Anything \cite{guo2024stereo}, most approaches typically utilize only one or two datasets, failing to fully leverage the existing stereo datasets.
\textit{Under-utilization of the diffusion architecture and priors.}
A diffusion foundation model for stereo depth estimation, similar to the Marigold model \cite{ke2024repurposing}, has not been explored.

\section{Multi-View Image Depth Estimation}
\label{sec:mulit-view_image}

\begin{figure}[ht]
  \centering
  \includegraphics[width=\linewidth]{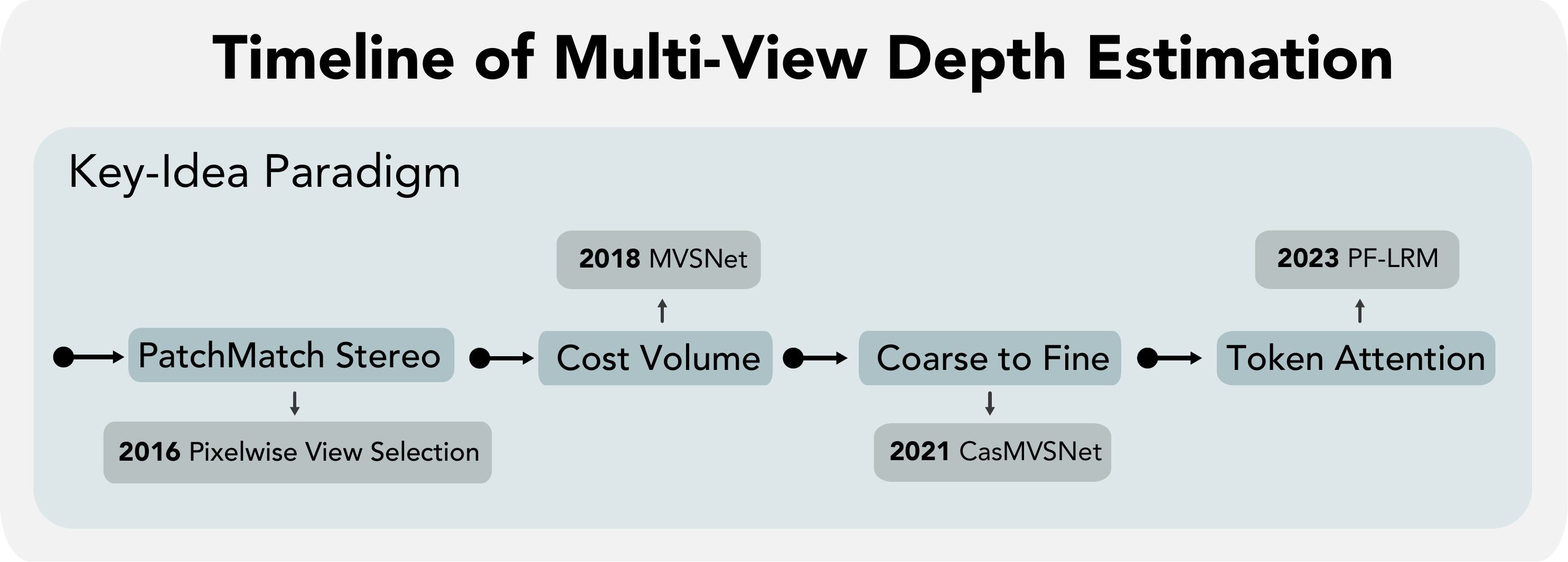}
  \caption{
    \textbf{Overview of the key-idea paradigm evolution of multi-view image depth estimation}.
    From early heuristic matching, through 2D CNNs and 3D CNNs, to advanced frameworks utilizing transformers and diffusion models, multi-view depth estimation models have shown increasing robustness and accuracy.
  }
  \label{fig:multiview_timeline}
\end{figure}

\begin{figure*}[ht]
  \centering
  \includegraphics[width=\linewidth]{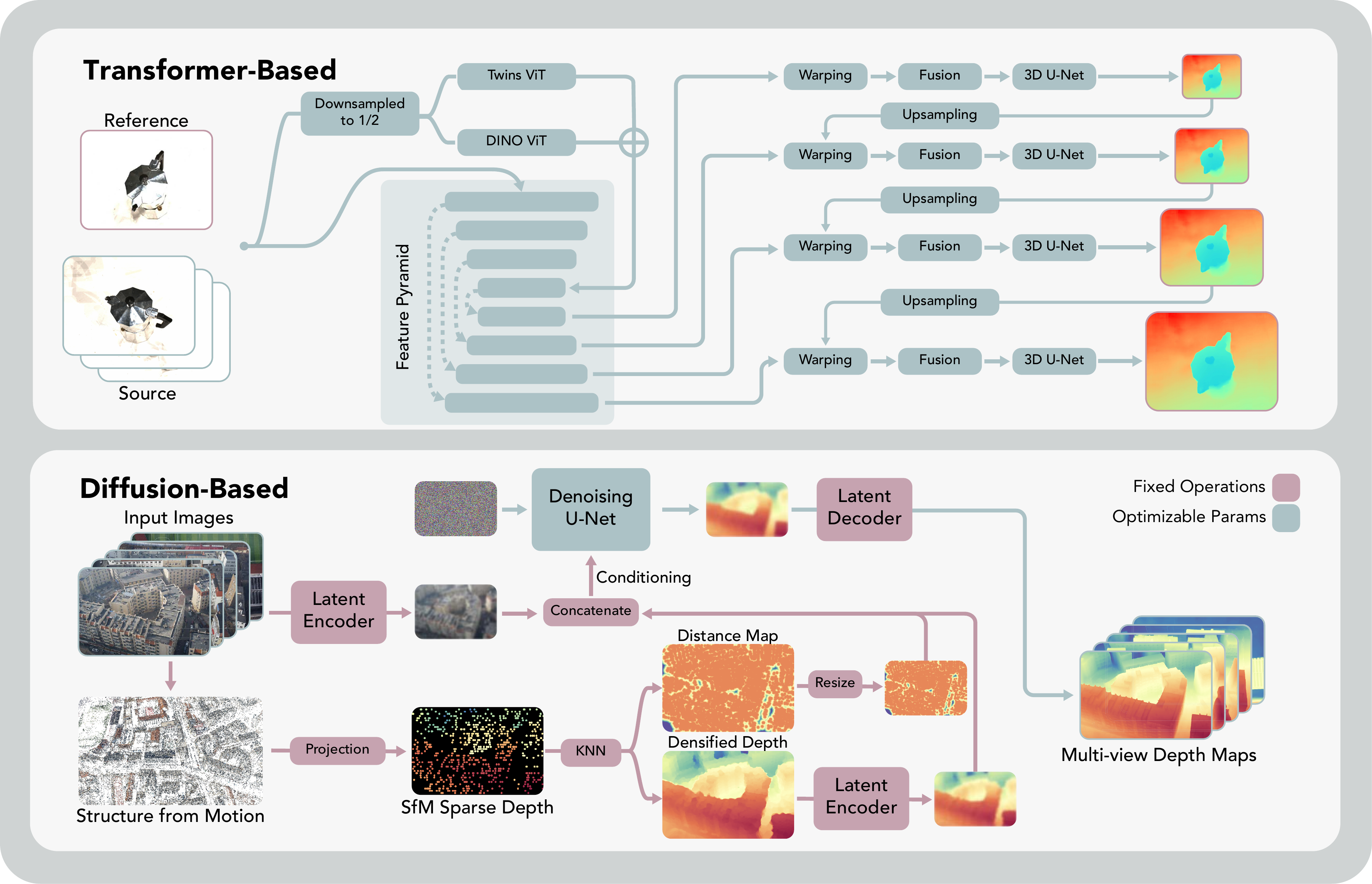}
  \caption{
    \textbf{Representative pipelines of multi-view image depth estimation}.
    As shown in the upper part of the figure, the Transformer-based architecture utilizes a Transformer model to extract global features and employs a 3D-geometry Transformer to facilitate cross-view information interaction (the Cost Volume method is depicted in the figure). In contrast, the Diffusion-based model in the lower part of the figure leverages the SfM (Structure from Motion) point clouds from multi-view images as prior information and generates depth maps through a diffusion model.
  }
  \label{fig:multiview_arch}
\end{figure*}

Multi-view image depth estimation (more commonly known as multi-view stereo, MVS) aims to recover the 3D structure of a scene from multiple images captured from different viewpoints, leveraging known or estimated camera poses. This task is fundamental to 3D reconstruction and underpins applications such as robotics, AR/VR, and autonomous navigation. Compared to monocular and stereo settings, the multi-view scenario provides richer geometric cues and enables more accurate and complete scene understanding, but also introduces new challenges in terms of view selection, feature aggregation, and computational efficiency. It's worth noting that the estimated depth maps are typically already aligned with the camera parameters from multiple viewpoints; thus, we do not distinguish between relative and metric depth estimation in this section. Over the years, the field has evolved from early heuristic and patch-based matching methods, through cost volume-based deep learning approaches with 2D and 3D CNNs, to recent advances utilizing transformers and diffusion models. These developments have not only improved the robustness and accuracy of depth estimation, for example, in sparse-view cases, but also paved the way for large-scale pretraining and the emergence of foundation models with strong generalization capabilities. In this section, we review the key technical paradigms, architectural innovations, and open challenges in multi-view image depth estimation, highlighting the trajectory towards foundation models and the remaining obstacles for real-world deployment.

\PAR{Evolution of model architectures.}
The model architectures for multi-view depth estimation have evolved from early heuristic matching to state-of-the-art Transformer and diffusion-based models.
Early approaches~\cite{bleyer2011patchmatch,colmap-mvs} typically assume that camera parameters were already known, obtained via SfM~\cite{colmap-sfm,triggs2000bundle,hartley2003multiple}, SLAM~\cite{montemerlo2002fastslam,davison2007monoslam}, or robotic arm control~\cite{tsai1989new,daniilidis1999hand}, and rely on traditional dense pixel matching techniques~\cite{barnes2009patchmatch} for depth estimation. This heuristic matching lays the groundwork for future developments.
With the advent of deep learning, researchers have begun employing 2D CNNs~\cite{mvdepthnet,yang2022mvs2d} to process the cost volume, which requires low computational cost and enables real-time scene reconstruction.
At the same time, alternative approaches utilize 3D CNNs~\cite{kar2017learning,ji2017surfacenet,mvsnet,huang2018deepmvs,ucsnet,casmvsnet,cvpmvsnet,vamvsnet,wang2022itermvs,xu2024igev++} to achieve more accurate depth estimation.
Recent trends have started to explore the use of Transformers~\cite{wang2022mvster,caomvsformer,cao2024mvsformer++,xu2024grm,wang2024vggsfm,wang2025vggt,mast3r_arxiv24,lu2025align3r,wang2025continuous,wang20243d,duisterhof2024mast3r,yifan2022input} or diffusion models~\cite{guo2025murre}, which leverage global self-attention mechanisms or diffusion-based strategies to capture richer feature representations and enhance inference quality.
\PAR{Evolution of method paradigms.}
The method paradigms for multi-view depth estimation have evolved from early PatchMatch-based matching strategies to modern token attention mechanisms.
Traditional methods~\cite{bleyer2011patchmatch,colmap-mvs} rely on the PatchMatch algorithm~\cite{barnes2009patchmatch} for pixel-level matching—a simple yet influential approach that sets the stage for subsequent innovations.
With the integration of deep learning, constructing a cost volume by back-projecting multi-view features using camera parameters becomes the mainstream~\cite{kar2017learning,ji2017surfacenet,mvsnet,mvdepthnet,huang2018deepmvs}, allowing for more accurate depth regression by computing feature correlations.
Given the high memory consumption of cost volumes, modern approaches~\cite{ucsnet,casmvsnet,cvpmvsnet,vamvsnet,wang2022itermvs,xu2024igev++} adopt a coarse-to-fine strategy: starting with a low resolution and a large number of depth hypotheses, and progressively refining the cost volume with predictions from previous stages to achieve a balance between high resolution and high accuracy.
Looking ahead, emerging research is investigating the incorporation of token attention~\cite{dust3r_cvpr24,xu2024grm,wang2025vggt,zhang2025flare,yang2025fast3r,jang2025pow3r,li2025megasam,shriram2024realmdreamer,zhang2024monst3r,mast3r_arxiv24,lu2025align3r,wang2025continuous,wang20243d,duisterhof2024mast3r} mechanisms in the depth estimation process, aiming to better capture both local and global contexts and offering promising new directions for multi-view depth estimation.

\PAR{Evolution towards foundation models.}
In terms of \textit{Model Size}, recent advances have been driven by the emergence of Transformer-based foundation models. MVSTR~\cite{wang2022mvster} is the first to introduce the Transformer architecture to multi-view depth estimation, employing a global-context Transformer and a 3D-geometry Transformer for intra-view global feature extraction and inter-view information interaction. MVSFormer~\cite{caomvsformer} proposes to use a pre-trained Vision Transformer (ViT) to enhance Multi-View Stereo (MVS) tasks, leveraging priors learned from large-scale datasets. MVSFormer++~\cite{cao2024mvsformer++} further employs a pre-trained DINOv2 model with 1.1 billion parameters and introduces distinct attention mechanisms tailored for the feature encoder and cost volume regularization. Both PF-LRM~\cite{wang2023pf} and DUSt3R~\cite{dust3r_cvpr24} utilize ViT to directly predict Point Maps, with DUSt3R employing the CroCo model featuring a complete encoder-decoder Transformer architecture.
In terms of \textit{Data Scale}, the evolution shows a clear trend from small-scale datasets to large-scale training. Early methods like MVSNet~\cite{mvsnet} used only 27K images from the DTU dataset, while MVSFormer++~\cite{cao2024mvsformer++} expanded to 40K images from the DTU and BlendedMVS datasets. The breakthrough came with sparse-view methods: PF-LRM~\cite{wang2023pf} utilizes 1M objects from Objaverse and MVImgNet datasets, DUSt3R~\cite{dust3r_cvpr24} is trained on 17M image pairs from eight datasets, including Habitat, MegaDepth, and Waymo, GRM~\cite{xu2024grm} uses 40M objects from Objaverse, and VGGT~\cite{wang2025vggt} employs 30M images from multiple datasets, including Co3Dv2, BlendMVS, and MegaDepth. These models establish a paradigm shift from traditional optimization-based methods to data-driven architectures capable of unified feature learning and cross-view reasoning.

\PAR{Valuable Problems.}
\textit{Sparse view reconstruction.}
In real-world applications, capturing a scene with dense and complete views may be feasible due to constraints on reachability. Therefore, the ability to reconstruct complete scene geometry from partial observations by leveraging prior knowledge represents an important direction for future research.
\textit{Find-grained depth estimation.}
Current feed-forward methods~\cite{wang2022mvster,caomvsformer,cao2024mvsformer++,xu2024grm} can efficiently predict multi-view image depth in a single forward pass. However, accurately capturing fine-grained geometry remains challenging, as it demands high-precision geometric prediction capabilities from neural networks.
\textit{Depth estimation of objects with complex materials.}
Reflective or transparent scenes pose significant challenges for geometry estimation due to their complex optical properties. Incorporating learned priors into depth estimation presents a promising approach for accurately capturing the geometry of complex materials.

\section{Monocular Video Depth Estimation}
\label{sec:monocular_video}

\begin{figure}[ht]
  \centering
  \includegraphics[width=\linewidth]{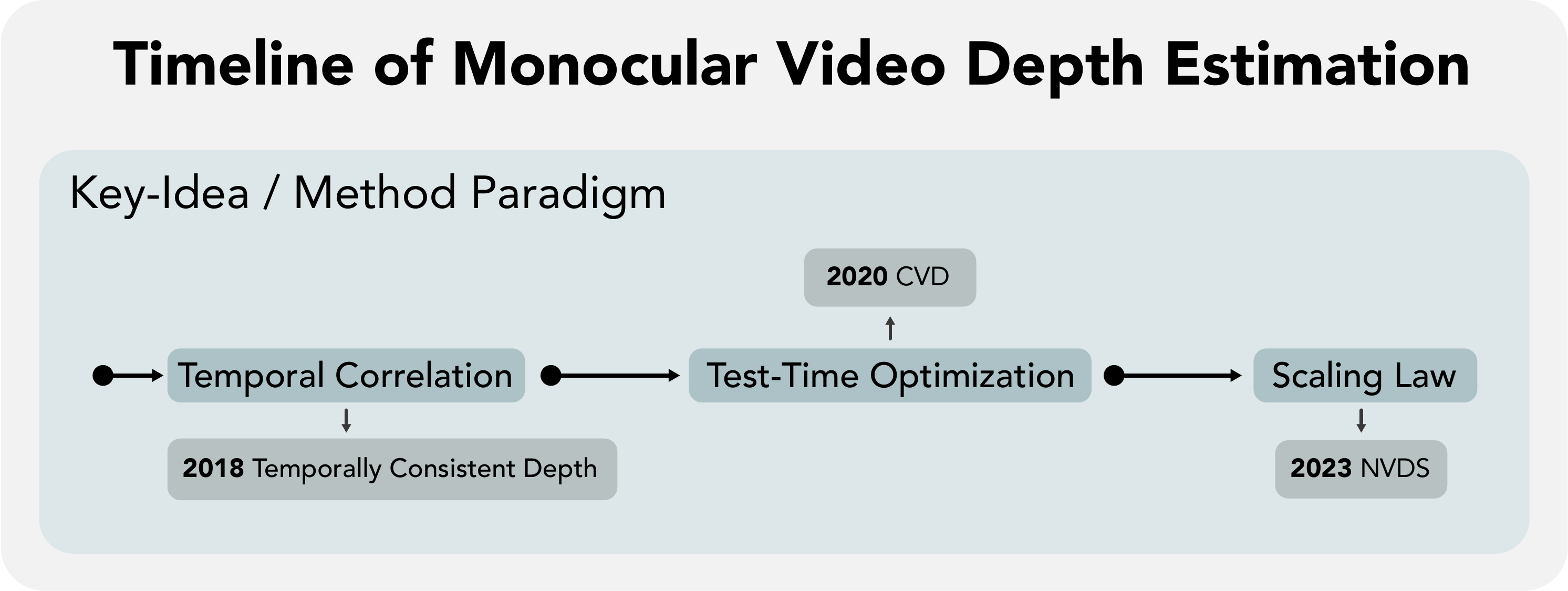}
  \caption{
    \textbf{Overview of the key-idea paradigm evolution of monocular video depth estimation}.
    From RNN-based temporal modeling (2019) to CNNs with test-time optimization (2020, CVD), transformer-based scaling (2023, NVDS), and video diffusion for enhanced stability and generalization.}
  \label{fig:video_timeline}
\end{figure}

Video depth estimation aims to estimate per-frame depth from a given monocular video while ensuring temporal consistency across the sequence. Compared to monocular depth estimation, its primary challenge lies in maintaining consistency over time. In contrast to multi-view stereo (MVS)~\cite{cao2024mvsformer++,xu2024grm} and other multi-view depth estimation methods~\cite{wang2022mvster,caomvsformer}, it needs to handle dynamic scenes, which imposes further challenges. Since video depth estimation requires both temporally consistent and accurate depth predictions, and its technical paradigm integrates elements from both monocular and multi-view depth estimation, we consider it the ultimate problem that a depth foundation model should address.
\PAR{Evolution of model architectures.} With the advancement of deep learning technologies and network architectures, various network structures have been applied in recent years to improve the temporal consistency of depth estimation.
With the rise and widespread adoption of RNNs in language models, in 2019,~\cite{zhang2019exploiting} proposed using the LSTM mechanism to integrate temporal information, thereby enhancing the stability of video depth estimation.
Meanwhile, in image tasks, CNN architectures (including U-Net and ResNet) have shown excellent performance. In 2020,~\cite{luo2020consistent} introduced test-time optimization to single-frame depth estimation methods based on CNN architectures, amplifying the capabilities of traditional CNNs through optimization and alignment.
Subsequently, attention mechanisms gained significant attention. ~\cite{yasarla2023mamo,wang2025continuous,lu2025align3r,wang20243d} applied attention mechanisms to integrate temporal features, maintaining memory through attention mechanisms rather than LSTMs, achieving more accurate and consistent depth estimation results.
Later, with the remarkable success of diffusion models in the field of image generation and the impressive results of video diffusion models, in 2024, ~\cite{shao2024learning,hu2024depthcrafter} proposed using video diffusion models to enhance consistency, achieving unprecedented stability and predictive performance.

\PAR{Evolution of method paradigms.} In recent years, innovative technical paradigms have been proposed to enhance the stability of video depth estimation.
In 2019,~\cite{zhang2019exploiting} introduced a memory mechanism to enable networks to integrate multi-frame information, implicitly encoding content from other frames to assist in depth estimation for the current frame. Subsequently, several works explored similar memory paradigms, such as MAMo~\cite{yasarla2023mamo}.
In 2020,~\cite{luo2020consistent} proposed using test-time optimization to post-process predicted depth results. Since this approach better leverages prior techniques like SfM and SLAM while also performing bundle adjustment (BA) on depth, it often yields superior results compared to purely generalized methods.
Later, with the rise of scaling laws, interest in generalized methods was reignited, and attention shifted to the generation and utilization of large-scale training data. In 2023,~\cite{wang2023neural} introduced a network-based post-processing method for monocular depth estimation results, finding a middle ground between direct video depth prediction and test-time training (TTT). It also introduced the first representative large-scale Video Depth in the Wild (VDW) dataset.
Video depth estimation has since begun to evolve toward the development of foundation models.

\begin{figure*}[ht]
  \centering
  \includegraphics[width=\linewidth]{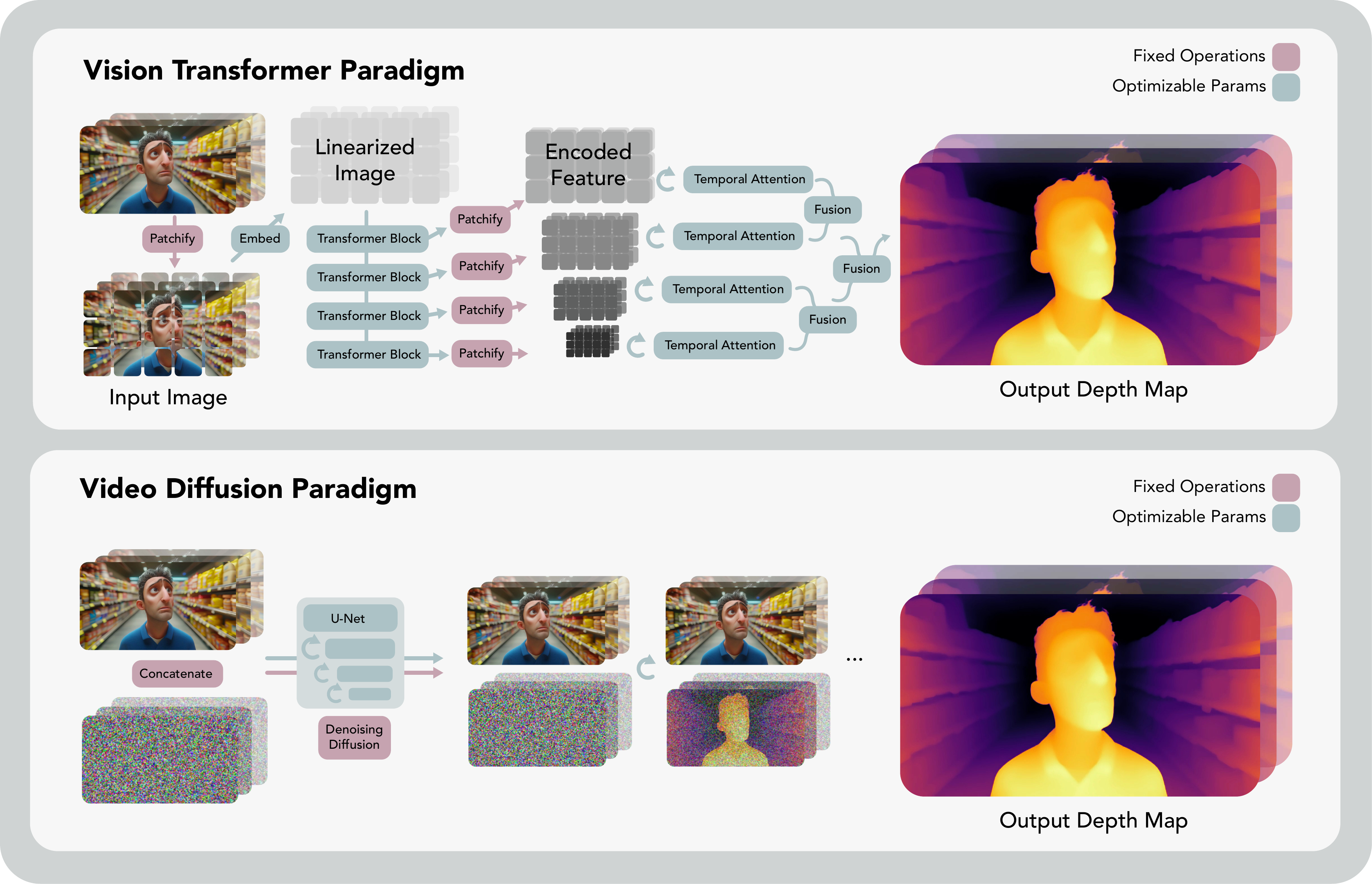}
  \caption{%
    \textbf{Representative pipelines of monocular video depth estimation}.
    The Vision Transformer (ViT) paradigm processes patchified inputs through self-attention and temporal attention mechanisms to integrate spatial and temporal depth cues. The Video Diffusion paradigm leverages denoising diffusion models, using concatenated depth and image features, optionally enriched with CLIP embeddings, to generate consistent video depth estimates. These scalable architectures enhance generalization and enable zero-shot depth estimation across diverse datasets.}
  \label{fig:video_arch}
  \vspace{-0.4cm}
\end{figure*}

\PAR{Evolution towards foundation models.}
In terms of \textit{Model Size}, recent work has focused on using scalable network architectures to model video depth. Buffer Anytime~\cite{kuang2024buffer} proposed using a Temporal ViT architecture with 343 million parameters to integrate temporal information and generate a large amount of pseudo ground truth (GT) based on single-image priors. Specifically, the input images are first patchified and converted into tokens recognizable by ViT. Then, self-attention is applied to these tokens. To fuse temporal information, cross-attention is performed on the single-frame attention results, reprocessing the features to obtain tokens integrated with temporal information. Finally, the tokens are decoded and unpatched to produce the final depth estimation results.
Another technical paradigm~\cite{shao2024learning,hu2024depthcrafter,yang2024depth} proposed using video diffusion models with 200 million to 1 billion parameters to aid in recovering video depth. Specifically, these methods use the input video as a condition in the video diffusion denoising process. For the noisy depth during denoising, it is concatenated with the corresponding frame's RGB color and, optionally, with CLIP embeddings as input to the denoising UNet.
In terms of \textit{Data scale}, the insufficient volume of video depth data has always been a critical issue \cite{wang2020dynocc}. In 2023, NVDS~\cite{wang2023neural} constructed the first large-scale, diverse video stereo depth dataset (14.2K videos, 2.24M frames) by using a video stereo matching method to generate pseudo labels. In 2024, methods like~\cite{shao2024learning,hu2024depthcrafter,yang2024depth} addressed the lack of video data by leveraging video priors from video generation models. Similar to NVDS~\cite{wang2023neural}, DepthCrafter~\cite{hu2024depthcrafter} annotated 200K videos using a video stereo matching method. Depth Any Video~\cite{yang2024depth} created 40K videos, totaling 6M frames, using a game engine. Video Depth Anything~\cite{chen2025video} proposed using an image teacher to provide pseudo labels to compensate for the shortage of video depth ground truth (GT) labels.

\PAR{Valuable problems.}
\textit{Geometric inconsistency.}
When camera motion is present in a monocular video, the ability to estimate consistent depth for the same statistical scene observed across multiple frames is crucial. To achieve this, jointly modeling camera motion and depth estimation is a promising approach and future direction for enhancing geometric consistency.
\textit{Temporal inconsistency.}
In the presence of dynamic objects, the lack of multi-view observation in monocular video makes it particularly challenging to estimate their geometry. The method needs to learn strong priors to predict temporally consistent depth for dynamic objects across multiple frames.
\textit{Monocular video depth training data.}
Learning strong dynamic priors needs extensive and diverse training data. However, due to the presence of dynamic objects, collecting real-world ground-truth monocular video training data with accurate ground-truth depth often requires additional depth sensors, limiting its scale-up capabilities.
Scaling up monocular video training data, either using real-world Internet unlabeled video with a self-supervised training strategy or simulating and rendering realistic synthetic data, is a valuable direction to be explored.

\section{Applications}
\label{sec:applications}

Depth foundation models are expected to see a wide range of practical applications across computer vision, robotics, and graphics.
By providing accurate, generalizable, and scalable depth estimation from monocular images, stereo pairs, multi-view images, or video sequences, these models could serve as a critical component for downstream tasks that require 3D scene understanding.
In this section, we highlight several representative application domains where depth foundation models would make a significant impact, including 3D reconstruction, novel view synthesis, video world modeling, and robotics and autonomous driving.

\subsection{3D Reconstruction}

For multi-view 3D reconstruction tasks, once the depth for each view is estimated, they can be directly fused to obtain the reconstructed scene geometry, which is usually represented as a point cloud or triangle mesh.
MVSNet~\cite{mvsnet} and CasMVSNet~\cite{casmvsnet} adopt fusibile~\cite{galliani2015massively}, which first converts the depth map of each view into a point cloud, then filters out points with poor consistency by projecting them into other views and checking, and finally fuses the point clouds from all views to obtain the reconstruction result of the entire scene.
Simplerecon~\cite{sormann2023simplerecon} and Murre~\cite{guo2025murre} use TSDF fusion~\cite{curless1996volumetric}, which converts depth maps into sparse truncated SDF grids, averages them across multiple views, and finally extracts the surface using marching cubes to obtain a triangle mesh.
Direct fusion methods are relatively efficient but require high-quality depth maps. NeuRIS~\cite{wang2022neuris} and MonoSDF~\cite{yu2022monosdf} use the relative depth predicted from a monocular depth estimation method as a supervision signal, constraining the SDF field through a specially designed depth loss, thereby enhancing the quality of the 3D reconstruction.

\subsection{Novel View Synthesis}
In recent years, NeRF-based \cite{mildenhall2021nerf} and 3D Gaussian Splatting-based \cite{kerbl3Dgaussians} methods have made significant progress in the task of novel view synthesis. High-quality depth estimations can serve as strong priors for these models, enhancing their performance and accelerating convergence. Specifically, the depth map can be utilized as sampling guidance for NeRF-based methods \cite{yu2022monosdf, guo2023streetsurf, miao2024efficient} and as a coarse initialization for 3D Gaussian Splatting-based methods \cite{fan2024instantsplat, zhou2024hugsim, miao2025evolsplat}. Additionally, the dense depth map can also be employed to learn a dense SDF field during the training process, allowing for the alignment of the geometry of NeRF or 3D Gaussians with the SDF field. This alignment can improve the geometry of the reconstructed results and facilitate the synthesis of high-quality novel views, particularly for perspectives that are distant from the training views.
Some existing works \cite{yan2024street, sandstrom2024splat, pan2025pings, matsuki2024gaussian, mao2024ngel} depend on lidar point clouds or RGB-D images as geometric priors. However, both lidar point clouds and RGB-D images can be costly to acquire, requiring additional sensors. High-quality depth estimations can serve as a substitute for these methods, unleashing the potential of novel view synthesis techniques.

\subsection{Video World Models}

As the popularity of diffusion models grows, video diffusion models have been proposed to generalize the image synthesis pipeline to video generation.
With the success of SORA \cite{sora2024} and other video foundation models \cite{hong2022cogvideo,yang2024cogvideox}, there exist several attempts \cite{rigter2024avid,zhang2024world,kang2024far} at exploring video models' capabilities as world models. Having a foundational model for depth estimation, preferably on videos, would significantly bridge the gap between image-only generation models' understanding of 3D space and motion. As the ability to predict depth would indicate, the model at least possesses the capability of distinguishing the size and placement pattern of everyday objects, scenes, and people. Having depth cues for video generation models could potentially serve as a breaking point for further boosting current world models' ability to understand everyday scenes and might even stimulate the underlying generalization ability even further, leading to more spatially and temporally consistent generation and future prediction results.

\subsection{Robotics and Autonomous Driving}
In robotics and autonomous driving, accurate and reliable depth perception plays a pivotal role in tasks such as navigation, obstacle detection, and collision avoidance.
Traditional solutions often rely on LiDAR or stereo camera systems, both of which come with increased hardware costs and complexity. Depth foundation models learned from large-scale datasets have the potential to deliver high-quality depth estimates from a single camera, making them particularly attractive for cost-sensitive real-world applications. Recent methods~\cite{lin2024promptda} demonstrate that monocular depth estimation can be integrated into robotic perception pipelines, serving as a complement or even a substitute for more expensive sensors.
For instance, some works \cite{yin2023metric3d,hu2024metric3d} adopt monocular depth estimation to enhance SLAM or visual odometry frameworks, showing improvements in localization and mapping under challenging lighting or weather conditions.
Similarly, in autonomous driving, depth estimates can facilitate large-scale 3D reconstructions for building realistic simulation environments, supporting algorithm development and testing~\cite{yan2024street}.
Furthermore, by leveraging depth priors learned from diverse scenes, these models exhibit promising generalization capabilities, potentially enabling robust domain adaptation across varied environments—from urban streets to off-road terrains—thus paving the way for more versatile and scalable robotic and self-driving solutions.

\section{Future Work}
\label{sec:future_work}
As discussed in the previous sections, there exist several fundamental problems to be solved before we reach a general-purpose depth foundation model, namely, data and consistency.

\PAR{Data.}
For all of the discussed depth estimation tasks, including monocular image, stereo image, multi-view image, and monocular video depth estimation, the lack of accurate, large-scale, high-quality, and high-variability data is currently the main concern for constructing and training a depth foundation model.
Due to the unique nature of the depth estimation task, current approaches to acquiring data usually fall into two categories: depth sensor or synthetic rendering.
For depth sensors, the main approach is to utilize LiDARs or ultrasonic devices. However, the acquired ground truth depth maps are usually incomplete or noisy due to the sensitive nature of the depth sensing devices.
For synthetic data generation, there exist several attempts at curating high-quality, hand-crafted, large static or dynamic 3D scenes by artists. However, these data are naturally limited to a small scale due to the amount of work required.
Future works should focus on either utilizing self-supervision techniques to better transfer the knowledge of vast image and video data to the task of depth estimation, or developing a better approach for simulation and generation, providing artist-quality synthetic rendering and depth pairs to boost generalization ability.

\PAR{Consistency.}
This includes both spatial and temporal consistency.
For the task of monocular image depth estimation, current methods typically fall short when merging depth estimation results together from different timestamps and viewports of the same scene.
For video depth estimation, although a vast amount of work has investigated the issue of temporal consistency throughout the target video, they still fail to produce accurate and consistent results when given multiple viewports of the same 3D scene or trying to unproject and merge the prediction results \cite{lin2024promptda}.
Notably, multi-view video reconstruction methods \cite{xu2024longvolcap,xu20234k4d,xu2023easyvolcap} have proved the existence of the dynamic and multi-view inductive bias of the 4D world by providing accurate reconstruction from only image-based optimization objects.
Future work should focus on exploring the intrinsic 3D or dynamic inductive bias present in the dynamic 3D world, further mitigating the problem of spatial and temporal inconsistency.

\section{Conclusion}
\label{sec:conclusion}

Since 2022, the advancements in foundation models within the natural language processing domain, along with the emergence of scaling laws, have led to a significant increase in the development of foundation models in the field of computer vision. In recent years, numerous foundation models have been introduced for depth estimation tasks, and new models continue to emerge at a rapid pace, making it challenging for practitioners to stay updated with the latest developments.

In this timely paper, we present a comprehensive survey of foundation models for depth estimation tasks, covering their background, development, and the latest advancements. We also address the valuable problems faced by existing depth foundation models and their downstream applications. We aim for this paper to serve as a valuable guide for practitioners and researchers interested in depth estimation foundation models.

Finally, there remain numerous challenges and opportunities in the realm of depth foundation models. We believe that as foundation models continue to evolve and depth estimation tasks advance, we will witness an increasing number of sophisticated and practical applications in the future.

\appendix

\subsection*{Availability of data and materials}

Not applicable.

\subsection*{Author contributions}
Zhen Xu and Hongyu Zhou were responsible for the overall writing of the manuscript.
Sida Peng, Yiyi Liao, Yue Wang, Ruizhen Hu, Xiaowei Zhou, and Hujun Bao provided critical supervision and guidance throughout the project, shaping its framework, refining technical discussions, and ensuring clarity and coherence, providing valuable oversight and feedback during drafting and revision.
The remaining co-authors supported the work by evaluating key publications and charting the evolution timeline of depth estimation architectures across the monocular, stereo, multi-view, and monocular video depth estimation tasks to deliver a thorough survey.

\subsection*{Acknowledgements}
This research was supported by Zhejiang Provincial Natural Science Foundation of China under Grant No. LD25F030001, and Information Technology Center and State Key Lab of CAD\&CG, Zhejiang University.

\subsection*{Declaration of competing interest}

This survey offers an analysis of recent vision-based depth estimation research and its trend towards depth foundation modeo, and does not introduce new datasets or materials, nor involve any competing interests.

\bibliographystyle{CVMbib}
\bibliography{main}

\subsection*{Author biography}

\begin{biography}[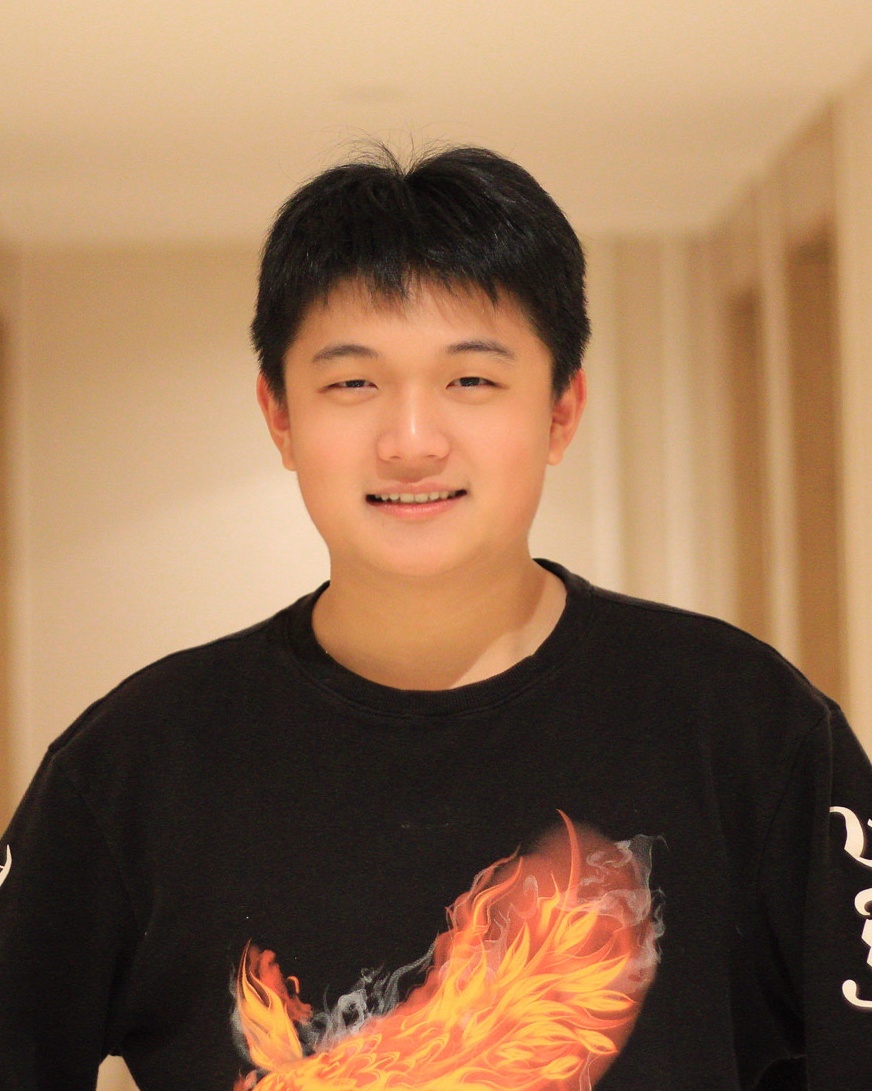]{Zhen Xu} is a forth-year PhD student in Computer Science at Zhejiang University, advised by Prof. Xiaowei Zhou and Sida Peng. He obtained his bachelor's degree in Computer Science from Zhejiang University in 2022. His current research focuses on 3D/4D neural reconstruction and rendering, and volumetric videos. 
\end{biography}

\begin{biography}[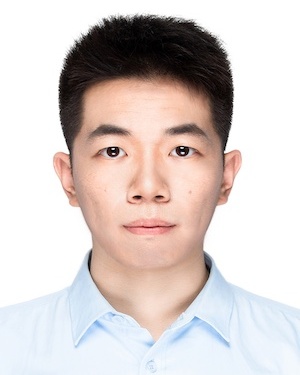]{Hongyu Zhou} is currently pursuing his PhD at Zhejiang University, working under the guidance of Prof. Yiyi Liao at the X-Dim Lab. Prior to this, he was a master student at Zhejiang University supervised by Prof. Deng Cai and Prof. Xiaofei at the State Key Lab of CAD\&CG. He received his bachelor’s degree from Tongji University in 2020.
\end{biography}

\begin{biography}[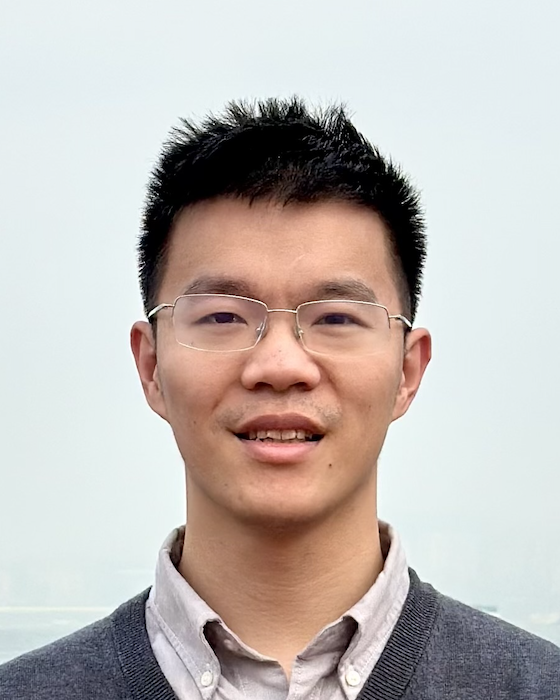]{Sida Peng} is an assistant professor (ZJU-100 Young Professor) at Zhejiang University. He received his Ph.D. degree from College of Computer Science and Technology at Zhejiang University in 2023, supervised by Prof. Xiaowei Zhou and Prof. Hujun Bao, and obtained his bachelor degree in Information Engineering from Zhejiang University in 2018. He received the 2024 CCF Outstanding Doctoral Dissertation Award and was selected as the 2022 Apple Scholar in AI/ML.
\end{biography}

\begin{biography}[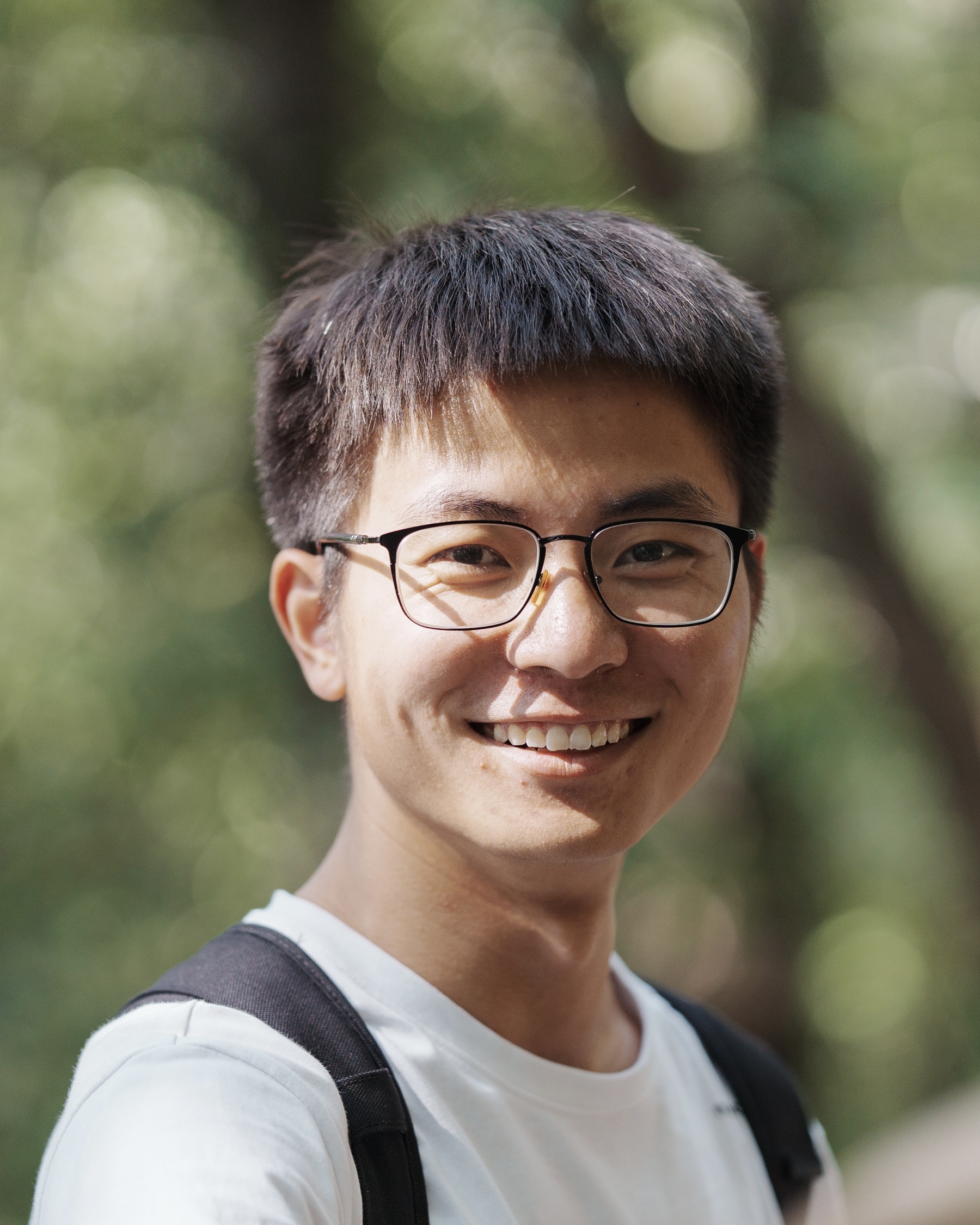]{Haotong Lin} is a fourth-year PhD student in Computer Science at Zhejiang University, advised by Prof. Xiaowei Zhou and Sida Peng. He obtained his bachelor degree in Computer Science from Zhejiang University in 2021. In the summer of 2022, he had the opportunity to engage in a highly enjoyable collaboration with Noah Snavely at Cornell University.
\end{biography}

\begin{biography}[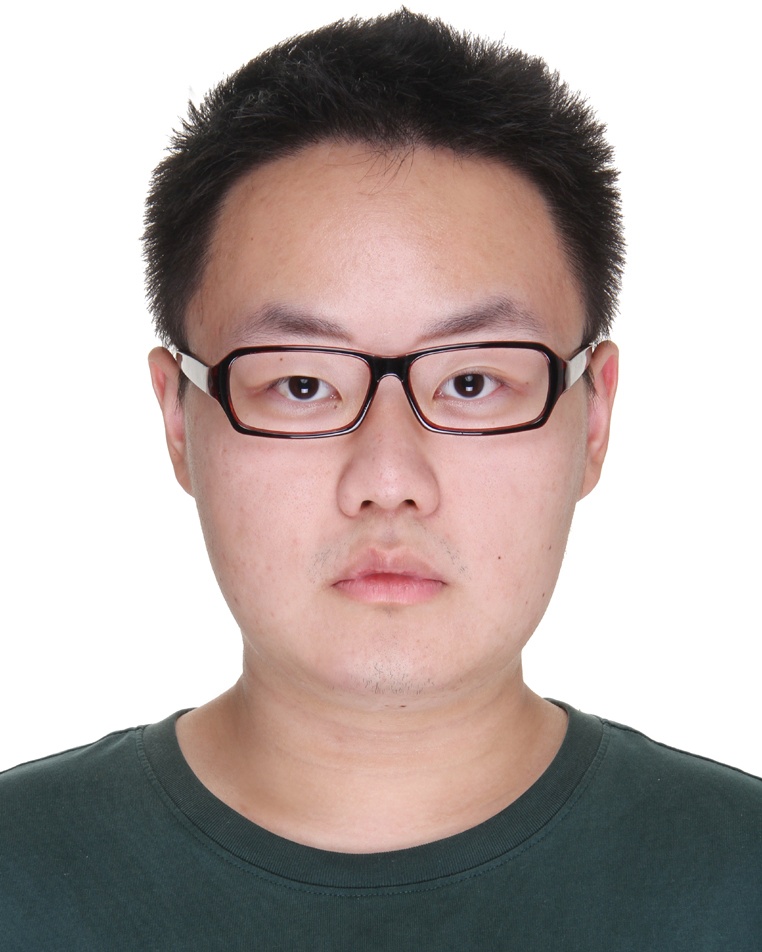]{Haoyu Guo} is a research scientist at Shanghai Artificial Intelligence Laboratory. He obtained his Ph.D. degree in Computer Science at Zhejiang University in 2025. His research interests include 3D vision, spatial intelligence and world model.
\end{biography}

\begin{biography}[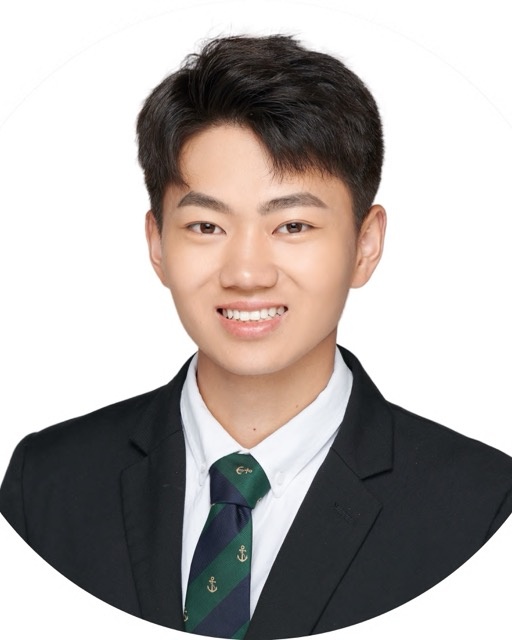]{Jiahao Shao} is a student researcher at X-D Lab advised by Yiyi Liao. He is also lucky to have collaboration with Matteo Poggi. Previously he obtained his B.Eng. degree in Automation from Zhejiang University in 2024.
\end{biography}

\begin{biography}[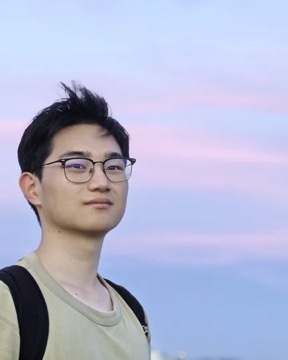]{Peishan Yang} is a Master's student of Computer Science at Zhejiang University (ZJU). His research interests focus on 3D reconstruction and scene understanding.
\end{biography}

\begin{biography}[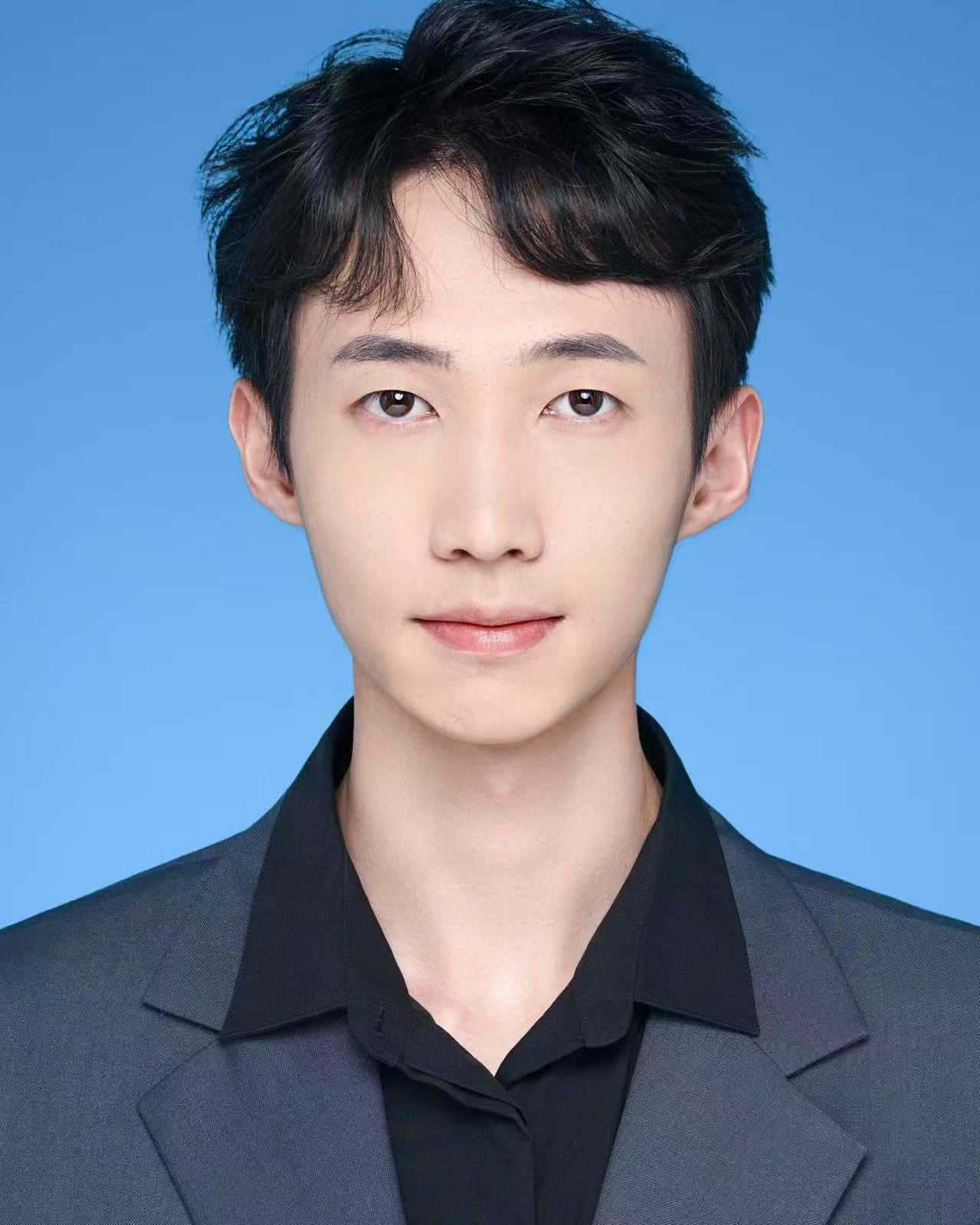]{Qinglin Yang} is currently a master’s student at the School of Software Engineering, Zhejiang University, advised by Dr. Sida Peng and Dr. Xiaowei Zhou. He received his B.E. degree in the School of Remote Sensing and Information Engineering, Wuhan University in 2024. His research interests include 3D foundation models for perception, understanding and generation.
\end{biography}

\begin{biography}[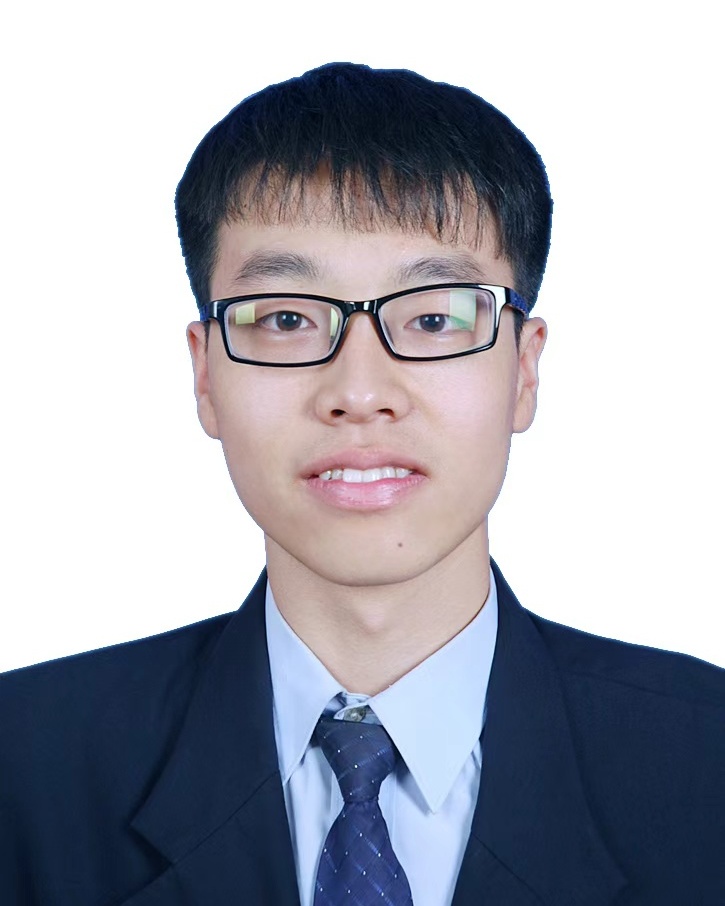]{Sheng Miao} is a PhD student at Zhejiang University. His research interests include 3D Reconstruction, Generative Model in autonomous driving.
\end{biography}

\begin{biography}[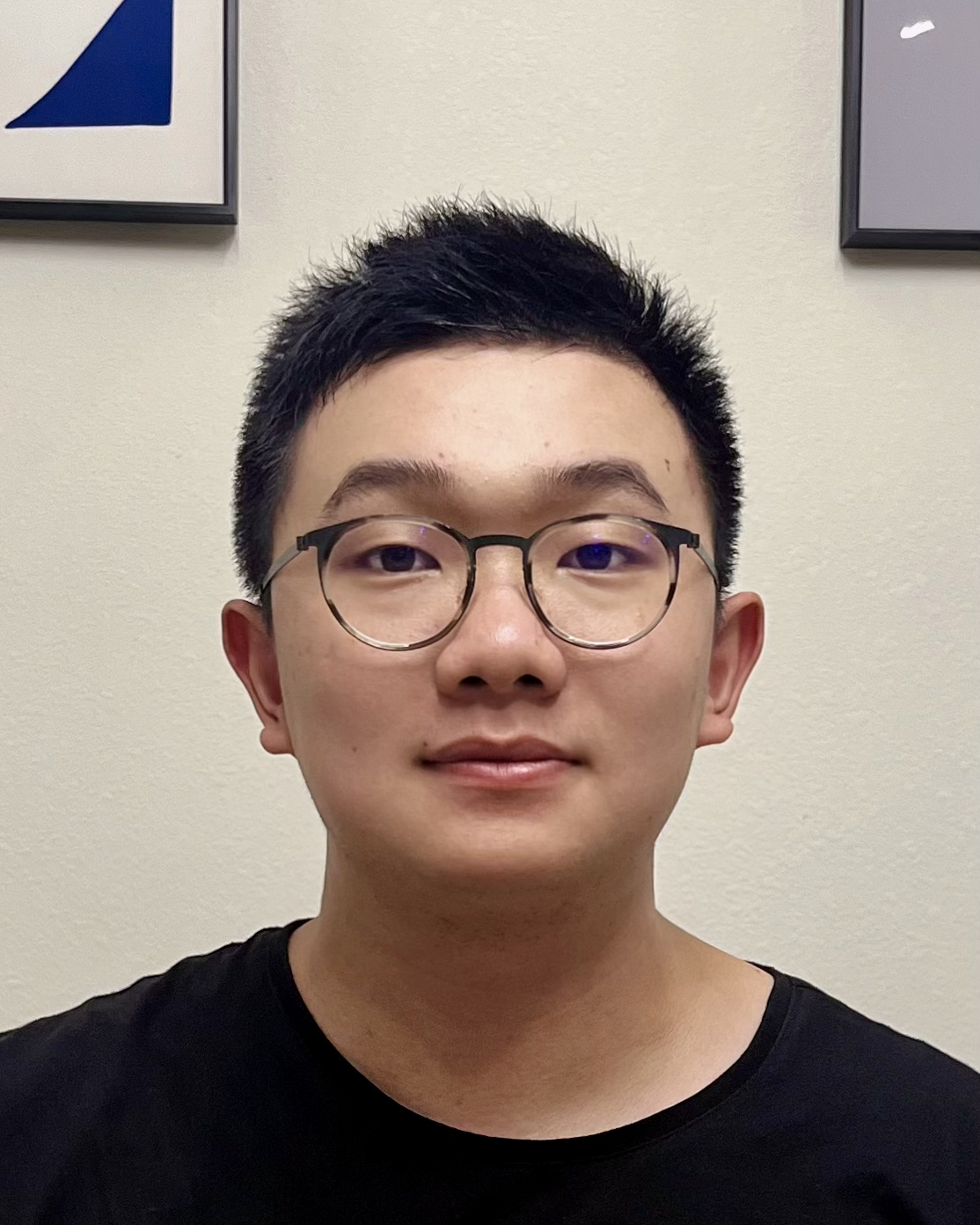]{Xingyi He} is a fourth-year Ph.D. student in Computer Science at Zhejiang University, advised by Prof. Xiaowei Zhou. He received his bachelor's degree from Huazhong University of Science and Technology in 2021. His research interests include 3D computer vision, robotics, image matching, and 3D reconstruction.
\end{biography}

\begin{biography}[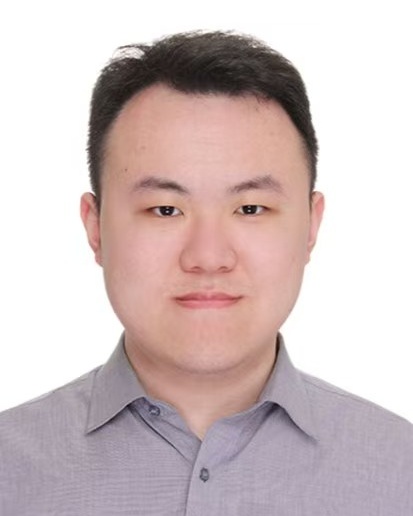]{Yifan Wang} is a MS student at Zhejiang University. His research interests include 3D vision, Dynamic Reconstruction, Neural Rendering.
\end{biography}

\begin{biography}[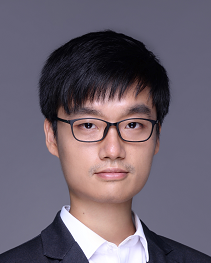]{Yue Wang} is a Professor in the Department of Control Science and Engineering at Zhejiang University. He received his Ph.D. from Zhejiang University in 2016, advised by Prof. Rong Xiong. His research focuses on persistent robot autonomy using AI, machine learning, and computer vision. He has developed various robotic systems, some of which have been applied in industry.
\end{biography}

\begin{biography}[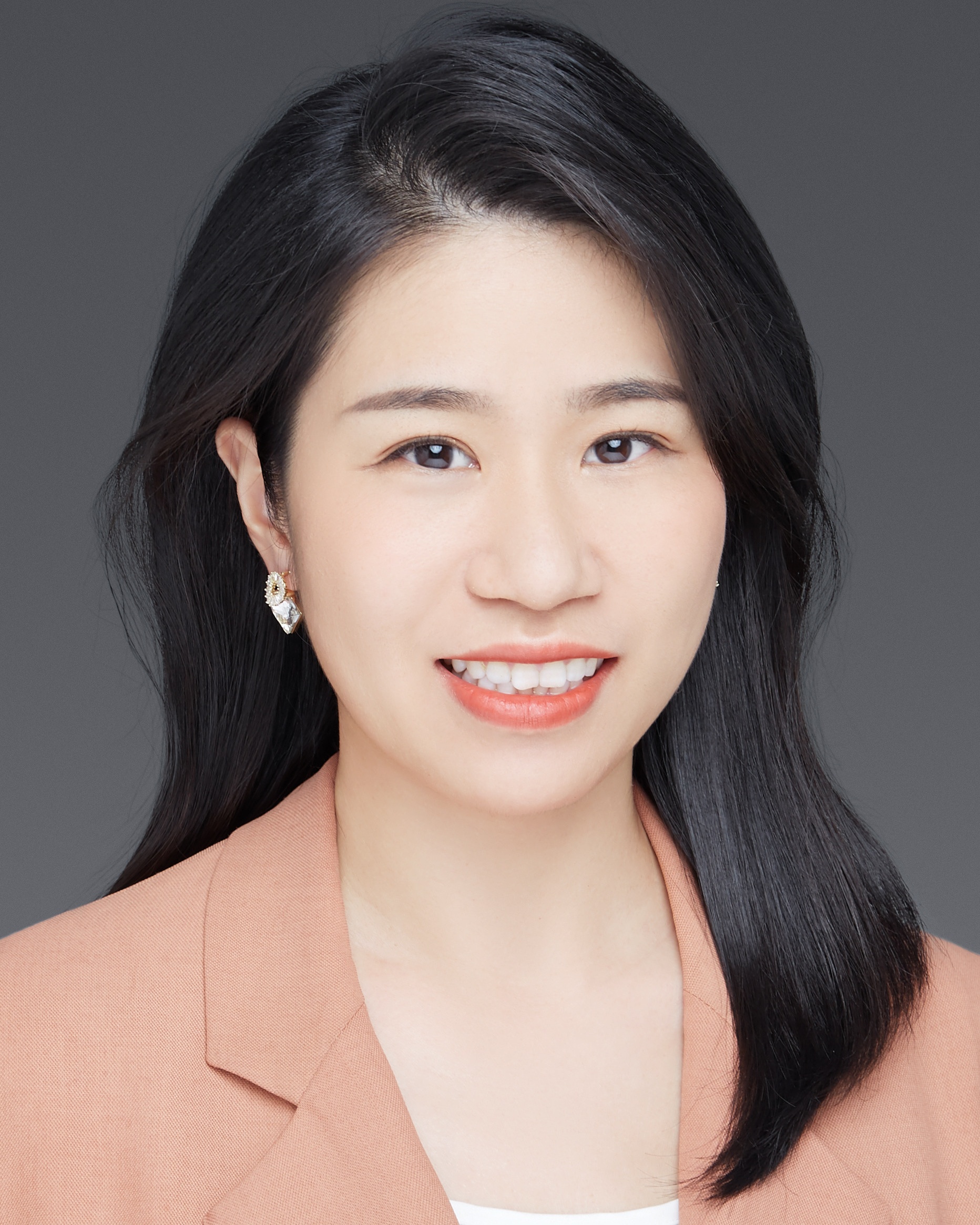]{Ruizhen Hu} is a Distinguished Professor and Deputy Director of the Visual Computing Research Center at Shenzhen University. She received her Ph.D. in Applied Mathematics from Zhejiang University in 2015, and was previously an Assistant Researcher at SIAT. Her research focuses on computer graphics and embodied AI, especially 3D modeling and agent interaction.
\end{biography}

\begin{biography}[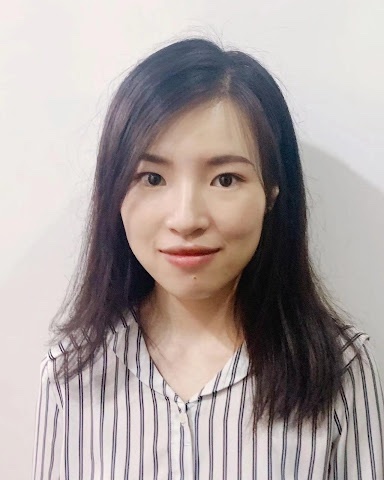]{Yiyi Liao} is an assistant professor at Zhejiang University, leading the X-D Lab. She was previously a postdoctoral researcher in the Autonomous Vision Group at the University of Tübingen and the Max Planck Institute for Intelligent Systems. She received her Ph.D. from Zhejiang University in 2018 and her B.S. from Xi’an Jiaotong University in 2013. Her research interests include 3D computer vision, scene understanding, and 3D generative models.
\end{biography}

\begin{biography}[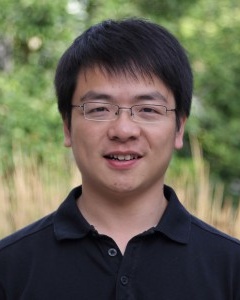]{Xiaowei Zhou} is a Professor in the College of Computer Science and the State Key Laboratory of CAD\&CG at Zhejiang University. Before joining ZJU in 2017, he was a postdoctoral researcher at the GRASP Lab, University of Pennsylvania. His research focuses on 3D computer vision, including 3D reconstruction, pose estimation, motion capture, scene understanding, and their applications in mixed reality and robotics.
\end{biography}

\begin{biography}[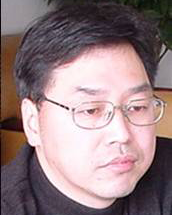]{Hujun Bao} is a Professor at the State Key Laboratory of CAD\&CG and the College of Computer Science and Technology, Zhejiang University. He received his B.Sc. (1987) and Ph.D. (1993) in mathematics and applied mathematics from Zhejiang University. He leads the 3D graphics computing group, focusing on geometry computing, 3D visual computing, and real-time rendering. His contributed to systems such as VisioniX, ACTS, and 2D-to-3D video conversion.
\end{biography}

\vspace*{2.6em}

\subsection*{Graphical abstract}
\begin{figure}[hb]
  \centering
  \includegraphics[width=\linewidth]{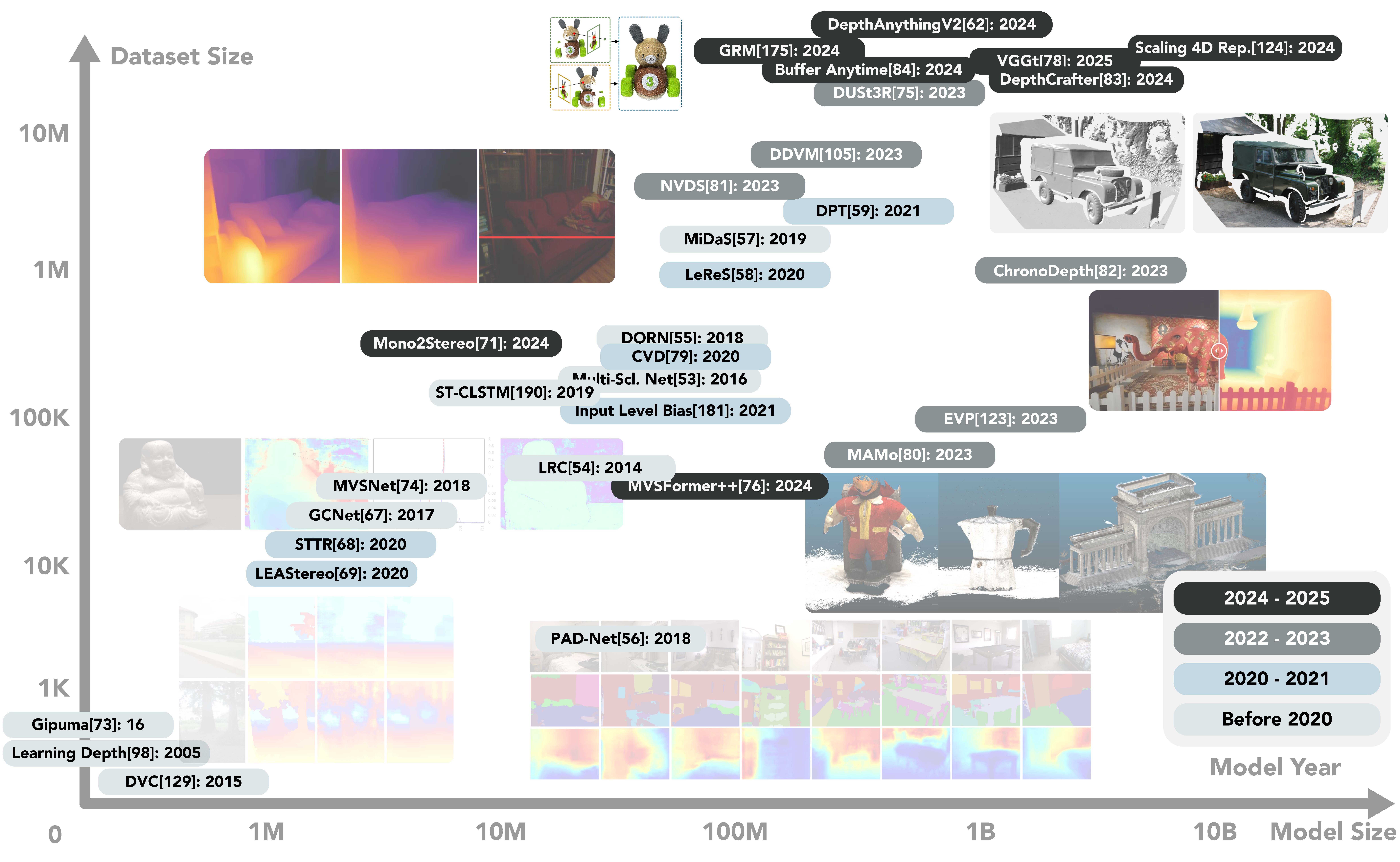}
\end{figure}

\end{document}